\documentclass[12pt]{article}

\usepackage{hyperref}

\usepackage{graphicx}
\RequirePackage{amsmath,mathrsfs,multirow,morefloats,booktabs,subfigure,pdfpages,color,algorithm,array,hhline,makecell,epstopdf,algorithm}
\RequirePackage[authoryear]{natbib}
\RequirePackage{algorithm,algpseudocode}

\numberwithin{equation}{section}
\newtheorem{thm}{Theorem}
\newtheorem{defin}[thm]{Definition}
\newtheorem{lem}[thm]{Lemma}

\begin{document}

	
	\title{\bf Community Detection by Principal Components Clustering Methods}
	
	\author{
		Huan Qing\thanks{Department of Mathematics,
			China University of Mining and Technology, China. }~
		 and
		Jingli Wang\thanks{Corresponding author. Email: jlwang@nankai.edu.cn. School of Statistics $\&$ Data Science,	Nankai University, China}
		
	}
	\date{}	
	\maketitle
%
	
	\begin{abstract}
		Based on the classical Degree Corrected Stochastic Blockmodel (DCSBM) model for network community detection problem, we propose two novel approaches: principal component clustering (PCC) and normalized principal component clustering (NPCC). Without any parameters to be estimated, the PCC method is simple to be implemented. Under mild conditions, we show that PCC yields consistent community detection. NPCC is designed based on the combination of the  PCC and the RSC method \citep{RSC}. Population analysis for NPCC shows that NPCC returns perfect clustering for the ideal case under DCSBM. PCC and NPCC is illustrated through synthetic and real-world datasets. Numerical results show that NPCC provides a significant improvement compare with PCC and RSC. Moreover, NPCC inherits nice properties of PCC and RSC such that NPCC is insensitive to the number of eigenvectors to be clustered and the choosing of the tuning parameter. When dealing with two weak signal networks Simmons and Caltech, by considering one more eigenvectors for clustering, we provide two refinements PCC+ and NPCC+ of PCC and NPCC, respectively. Both two refinements algorithms provide improvement performances compared with their original algorithms. Especially, NPCC+ provides satisfactory performances on Simmons and Caltech, with error rates of 121/1137 and 96/590, respectively.
	\end{abstract}
	
	\textbf{keyword:}
		principal component clustering; community detection; degree-corrected stochastic blockmodel; asymmetric regularized Laplacian matrix; weak signal networks
		
		MSC[2010]: 62H30;  91C20



\section{Introduction}
Community detection is a problem that has received considerable attention in past few decades in computer science \citep{RSCtau, RSC}, statistics \citep{PJBAC,CMM,SCORE}, social science \citep{ MN2006, nguyen2014dynamic} and biology \citep{liu2018global,tomasoni2020monet}. To find communities in networks and study the estimation properties, many authors have constructed a number of algorithms under various models.  \cite{fortunato2016community} and \cite{newman2004detecting} have provided some nice review of some historical approaches in various areas.  The stochastic block model (SBM) \citep{SBM} is a benchmark model for community detection.  Substantial approaches such as \cite{amini2013pseudo, PJBAC,daudin2008a,  nowicki2001estimation,  rohe2011spectral,snijders1997estimation} are designed based on SBM, since it is mathematically simple and relatively easy to analyze \citep{PJBAC}.  Though the versatility and analytic tractability of SBM has made it arguably the most popular model for studying community detection \citep{CMM}, it fails to detect real-world networks with heterogeneous nodes degrees. For example, as studied in \cite{SCORE}, two SBM-based methods oPCA and nPCA \citep{nPCA} fail to detect political blog network \citep{Polblogs1} with vary high error rates. The shortcoming of SBM comes from the unrealistic assumption that assumes nodes in the same cluster share the same expectation degree. To overcome this challenge, the degree-corrected stochastic block model (DCSBM) \citep{DCSBM} is proposed by introducing one degree heterogeneity parameter for each node. DCSBM is more realistic and accurate than SBM for real-world networks with high degree heterogeneity.

Based on DCSBM, an increasing number of community detection methods have been proposed. There are two popular directions to develop algorithms for community detection: model-based methods and spectral methods \citep{CMM}.  Model-based methods always intend to solve an optimization problem to seek for clusters, hence they are always computationally intractable. For example, \cite{MN2006} constructed a highly effective approach by optimizing a quality function (modularity) over the possible divisions of a network. A convexified modularity maximization (CMM) approach is established to estimate the hidden communities under DCSBM \citep{CMM}.  As a comparison,  spectral methods are computationally fast since they estimate the communities based on the eigenvectors of the network adjacency matrix and its variants. \cite{SCORE} proposed a consistent spectral clustering method on ratios-of-eigenvectors (named SCORE). \cite{OCCAM} developed an overlapping continuous community assignment model (OCCAM) for some sparse networks with overlaps. The regularized spectral clustering (RSC) \citep{RSC} is also a spectral methods which is based on a regularized Laplacian matrix. Some nice applications and discussions for spectral methods are provided in \cite{Tutorial} and \cite{ng2001on}. In this paper, we first design a computationally feasible spectral clustering algorithm which has strong statistical performance guarantees under DCSBM and enjoys satisfactory empirical performance. And then we also construct an improved version of the proposed method. 

In this paper, we adopt a clustering method for undirected and unweighted network with known clusters $K$. We first construct a dummy matrix which is defined as a production of eigenvalues and its eigenvectors of the adjacency matrix of the network. Then we normalize the dummy matrix by row. Finally, the labels for $K$ clusters are estimated by implying the method of k-means to the normalized matrix. This method is named Principal Component Clustering (PCC) method. The PCC is a concise method since it does not require any regularizer, threshold or tuning parameters. What's more, we build a theoretical framework for PCC and prove the consistency of estimation. Furthermore, motivated by \cite{RSCtau} and \cite{RSC}, we combine PCC and a regularized Laplacian matrix to improve the performance of PCC. This improved version of PCC is called Normalized Principal Component Clustering (NPCC). The NPCC is novel since it is designed based on the asymmetric matrix $N^{L_{\tau}}$ (given in the Definition \ref{defN}).  Naturally, NPCC inherits some nice properties of PCC and RSC. Furthermore, by borrowing the idea from SCORE+ method \citep{SCORE+}, our proposed approaches PCC and NPCC can also imply more eigenvectors and eigenvalues for clustering to deal with some weak signal networks. For simplification, we only present the algorithms PCC+ and NPCC+ in real data anlaysis.

This paper provides with three main contributions. The first one is that we design two spectral clustering methods PCC and NPCC with better numerical performances compared with traditional spectral clustering methods. PCC and NPCC are designed based on the idea that  clustering the matrix computed by the production of eigenvalues and eigenvectors of adjacency matrix or its variants of networks,  where this idea may inspire people to develop more spectral clustering methods. The second one is that our proposed approaches, especially NPCC which is designed based on an asymmetric matrix, can identify clusters with outstanding performances. The third one is that our methods PCC and NPCC can detect weak signal networks by considering one more eigenvector for clustering while traditional spectral clustering methods fail to detect weak signal networks. And we name the two refinements procedures as PCC+ and NPCC+, where PCC+ and NPCC+ can be generally applied to detect both strong and weak signal networks.

The rest of the article is arranged as follows. In section \ref{sec2}, we formulate the problem of community detection and review the model, DCSBM. In section \ref{sec:PCC}, we first propose a Principal Component Clustering (PCC) algorithm for community detection in section \ref{algPCC} and then study the theoretical properties of PCC in section \ref{popPCC} and \ref{HammingPCC}. Furthermore, we also construct a Normalized Principal Component Clustering (NPCC) in section \ref{sec:NPCC}. Specifically, we introduce the algorithm of NPCC in section \ref{algNPCC} and present some theoretical results in section \ref{popNPCC} and \ref{spectralNPCC}.  In section \ref{dataanalysis}, we evaluate the proposed clustering methods PCC and NPCC through extensive simulation studies and real-word networks, and we briefly present two algorithms PCC+ and NPCC+ in this section.  Section \ref{sec6} concludes this paper with discussions.


\section{Problem Formulation and Review of DCSBM}\label{sec2}
In this section, we set up the problem of community detection and briefly review the classical degree-corrected stochastic block model (DCSBM).

Consider an undirected and unweighted network $G(V,E)$, where $V=\{v_{1},v_{2},\ldots, v_{n}\}$ is the node set, $n$ is the number of nodes in the network and  $E$ is the set of edges such that $E$ contains a pair $(i,j)$ if there is an edge for any two distinct nodes $i,j$. The edge set can be represented by a symmetric adjacency matrix $A\in \{0,1\}^{n\times n}$ which is defined by
\begin{equation*}
A_{i,j}=
\begin{cases}
  1,&\mathrm{if~there~is~an~edge~between~node~}i\mathrm{~and~}j,\\
  0, & \mathrm{otherwise}.
\end{cases}
\end{equation*}
 Noted that the diagonal elements of $A$ are 0 (which means there is no self-connection).  Assume that the network is connected and with $K$ perceivable non-overlapping communities $V^{(1)}, V^{(2)},\ldots,V^{(K)}$, where the non-overlapping means that each node belongs to \emph{exactly} one cluster. $K$ is assumed to be known in this paper. The community labels $\ell =  (\ell(1), \dots, \ell(n))^{\prime}$ is a $n\times 1$ vector, where $\ell(i)$  denotes the community that node $i$ belongs to, for $i \in \{1,2,\ldots, K\}$.  $\ell$ is unknown and the goal of community detection is to use $(A,K)$ to predict nodes labels $\ell$.

The following notations will be used throughout the paper: $\|\cdot\|_{F}$ denotes the Frobenius norm for any matrix.  $\|\cdot\|$ denotes the spectral norm for any matrix. $\|\cdot\|$ for a vector denotes the $l_{2}$-norm. $\|\cdot\|_{3}$ for a vector denotes the $l_{3}$-norm.  For any matrix $M$, when we say ``leading eigenvalues'', we are comparing the magnitudes of the eigenvalues of $M$. When we say ``leading eigenvectors'', the leading eigenvectors are the leading eigenvalues' respective right unit-norm eigenvectors of $M$.
\subsection{The degree-corrected stochastic blockmodel (DCSBM)}
This paper is based on the degree-corrected stochastic blockmodel. We first review DCSBM in this subsection. Let $P$ be a $K\times K$ symmetric matrix whose elements are in $[0,1]$. Let $\theta$ be a $n\times 1$ vector such that $\theta_{i}$ is the $i$-th node's degree heterogeneity. In fact, $\theta_{i}$ is used to control the degrees of node $i$.  Under DCSBM, the probability that there is an edge between node $i$ and node $j$ is defined as $\theta_{i}\theta_{j}P_{g_{i}g_{j}}$, where $g_{i}$ denotes the cluster that node $i$ belongs to. Note that in the DCSBM model, if $\theta_{i}=\theta_{j}$ holds for all $i,j$, the DCSBM degenerates to the SBM model. 
In this paper, we assume that the  $K\times K$ mixing matrix $P$ should satisfy
\begin{align}\label{defP}
  P \mathrm{~is~symmetric},~\mathrm{nonnegative},~\mathrm{nonsingular~and~irreducible}.
\end{align}
Same as \cite{SCORE}, the adjacency matrix $A$ under DCSBM can be written as follows
\begin{align}\label{defA}
  A=E[A]+W=\Omega-\mathrm{diag}(\Omega)+W,
\end{align}
where $E[A]$ means the expectation of $A$ (in the DCSBM model, the expectation of $A(i,j)$ is $\theta_{i}\theta_{j}P_{g_{i}g_{j}}$, therefore given the $n\times 1$ degree heterogeneity  parameter vector $\theta$ and the $K\times K$ mixing matrix $P$, we can easily obtain $E[A]$). Set $W=A-E[A]$, then we can simply consider $W$ as the difference between the adjacency matrix $A$ and its expectation matrix $E[A]$. In (\ref{defA}), $\mathrm{diag}(\Omega)$ is a diagonal matrix whose diagonal elements are the diagonal elements of $\Omega$. And we treat $\Omega$ as the expectation matrix of $A$ just as in \cite{SCORE} and \cite{RSC} such that $\Omega_{ij}=\mathrm{Pr}(A_{ij}=1)=\theta_{i}\theta_{j}P_{g_{i}g_{j}}$. Then under DCSBM, the expectation of the adjacency matrix can be expressed as
\begin{align}\label{defOmega}
  \Omega=\Theta ZPZ'\Theta,
\end{align}
where (1) $\Theta$ is a $n\times n$ diagonal matrix whose $i$-th diagonal element is $\theta_{i}$ and (2)
$Z\in \{0,1\}^{n\times K}$ is the membership matrix with $Z_{ik}=1$ if and only if node $i$ belongs to block $t$ (i.e., $g_{i}=k$).

Model (\ref{defP})-(\ref{defOmega}) contribute to the K-community degree-corrected stochastic block model (DCSBM), and denote DCSBM by $DCSBM(n, P, \Theta, Z)$.
For the community detection problem in this paper, we aim at designing an efficient algorithm via analyzing the properties of $\Omega$ and its variants based on $DCSBM(n, P, \Theta, Z)$.
Note that though $\Omega$ is a $n\times n$ symmetric matrix, since we assume that $P$ should satisfy conditions in (\ref{defP}), by basic algebra, we know that the rank of $\Omega$ is $K$, hence $\Omega$ has $K$ nonzero eigenvalues.
\section{Principal Component Clustering (PCC)}\label{sec:PCC}
In this section we first introduce our Principal Component Clustering (PCC) method, and then present the theoretical results of PCC.
\subsection{The Algorithm of PCC}\label{algPCC} We call the algorithm as principal component clustering since this approach only applies the leading $K$ eigenvalues and respective unit-norm eigenvectors of $A$ for community detection. The algorithm of proposed PCC can be presented as follows:
\begin{algorithm}
	\caption{\textbf{Principal Component Clustering algorithm} (\textbf{PCC})}
	\label{alg:PCC}
	\begin{algorithmic}[1]
		\Require $A,K$
		\Ensure node labels $\hat{\ell}$
		\State Compute $\hat{X}=V_{A}E_{A}$ where $E_{A}$ is an $K\times K$ diagonal matrix that contains the leading $K$ eigenvalues of $A$, and $V_{A}$ is an $n\times K$ matrix contains the respective unit norm eigenvector of $A$ as column.
		\State  Compute $\hat{X}^{*}$ by normalizing each of $\hat{X}$' rows to have unit length, i.e., $\check{X}_{i}^{*}=\frac{\check{X}_{i}}{\|\check{X}_{i}\|_{2}}$, for i = 1, \dots, n.
		\State Apply k-means to $\hat{X}^{*}$ assuming there are $K$ clusters to obtain $\hat{\ell}$.
	\end{algorithmic}
\end{algorithm}

From the algorithm of PCC, we can find that PCC is pithy since it  does not require any extra parameters to be estimated, and it is easy to be applied for community detection. Meanwhile, unlike the two spectral clustering approaches RSC and SCORE, PCC take advantage of the leading $K$ eigenvalues of $A$ to re-weight $\hat{X}$, that's the reason that PCC is so pithy, it still has satisfactory numerical performances, just as shown in Section \ref{dataanalysis}, the numerical study part.
\subsection{Population analysis of the PCC}\label{popPCC}
Under DCSBM, if the partition is identifiable, one should be able to determine the partition from $\Omega$. This section shows that PCC perfectly reconstructs the block partition via the following population analysis.

Under $DCSBM(n, P, \Theta, Z)$, we have $\Omega=\Theta ZPZ'\Theta$, $\mathrm{rank}(\Omega)=K$, and $\Omega$ has $K$ nonzero eigenvalues, marked them as $\lambda_{1}, \lambda_{2}, \ldots, \lambda_{K}$ in descending order in  magnitude. Let $E_{\Omega}$ be the $K\times K$ diagonal matrix whose $i$-th diagonal entry is $\lambda_{i}$, and let $V_{\Omega}$ be the $n\times K$ matrix whose $i$-th column is $\eta_{i}$ for $1\leq i\leq K$, where $\eta_{i}$ is the $i$-th leading eigenvector of $\Omega$. Then define $X$ as $X=V_{\Omega}E_{\Omega}$ and define $X^{*}$ by normalizing each of $X$' rows to have unit length. To obtain a explicit form of $X^{*}$, we borrow the following Lemma \ref{lemma2.1Jin} from \cite{SCORE} (which is the Lemma 2.1 in \cite{SCORE}) since it presents the forms for leading $K$ eigenvectors of $\Omega$.
For the convenient of introducing Lemma \ref{lemma2.1Jin}, we introduce some notations. For $1\leq i\leq n, 1\leq k\leq K$, let $\theta^{(k)}$ be the $n\times 1$ vector such that $\theta^{(k)}(i)=\theta_i,$  if $g_{i}=k,$ and $\theta^{(k)}(i)=0,$ otherwise. Let $\bar{D}$ be the $K\times K$ diagonal matrix  such that $\bar{D}(k,k)=\|\theta^{(k)}\|/\|\theta\|, 1\leq k\leq K.$

\begin{lem}\label{lemma2.1Jin}
	 Under $DCSBM(n,P,\Theta,Z)$, suppose all eigenvalues of $\bar{D}P\bar{D}$ are simple. Let $\{\frac{\lambda_{i}}{\|\theta\|^{2}}\}_{i=1}^{K}$ be such eigenvalues arranged in the descending order of the magnitudes, and let $a_{1}, a_{2}, \ldots, a_{K}$ be the associated (unit-norm) eigenvectors. Then the $K$ nonzero eigenvalues of $\Omega$ are $\lambda_{1}, \lambda_{2}, \ldots, \lambda_{K}$, with the associated (unit-norm) eigenvectors being
	\begin{align*} \eta_{k}=\sum_{i=1}^{K}[a_{k}(i)/\|\theta^{(i)}\|]\cdot \theta^{(i)},\quad k=1,2,\ldots,K.
	\end{align*}
\end{lem}
The proof of Lemma \ref{lemma2.1Jin} can be find in \cite{SCORE}, and we omit it here. Recall that $X=V_{\Omega}E_{\Omega}$, sicne Lemma \ref{lemma2.1Jin} gives the explicit expressions of the leading $K$ eigenvectors of $\Omega$, now we find the explicit expressions of every entry of $X$, and hence the explicit form for $X^{*}$. Based on the known expression of $X^{*}$, the following lemma guarantees that applying k-means on $X^{*}$ returns perfect clustering, i.e., PCC returns perfect clustering under the ideal case. Lemma \ref{LemmaPCC} is shown in Appendix.
\begin{lem}\label{LemmaPCC}
	Under the $DCSBM(n, P, \Theta, Z)$, $X^{*}$ has exactly $K$ distinct rows, and if $g_{i}=g_{j}$ for any node $i\neq j$, then the $i$-th  row of $X^{*}$ equals to the $j$-th row of $X^{*}$.
\end{lem}
Lemma \ref{LemmaPCC} finishes the population analysis and guarantees that if we use $\Omega$ to replace $A$ in Algorithm \ref{alg:PCC}, then we can obtain the true nodes labels, suggesting that PCC returns perfect clustering results for the ideal case under $DCSBM(n, P, \Theta, Z)$.
\subsection{Bound of Hamming error rate of the PCC}\label{HammingPCC}
Hamming error rate can be used to measure the performance of clustering methods \cite{SCORE}. To find the bound for Hamming error rate, we first show that $\hat{X}^{*}$ is close to $X^{*}$ under certain conditions in Theorem \ref{mainPCC1}, and then give the bound of the Hamming error rate of the PCC method under DCSBM in Theorem \ref{mainPCC2}.

We make the following assumptions\footnote{Actually, the four assumptions (A1)-(A4) are the same as the four assumptions (2.9), (2.12), (2.13) and (2.15) in \cite{SCORE}, respectively. Because  \cite{SCORE} gives the two bounds of $|\hat{\lambda}_{k}-\lambda|$ and $\|\hat{\eta}_{k}-\eta_{k}\|$ for $1\leq k\leq K$, and the two bounds will be applied to bound $\|\hat{X}^{*}-X^{*}\|_{F}$ for our PCC in this paper, therefore this paper shares same assumptions as \cite{SCORE}.}:

(A1) $\mathrm{log}(n)\theta_{\mathrm{max}}\|\theta\|_{1}/\|\theta\|^{4}\rightarrow 0, \mathrm{as~} n\rightarrow \infty.$

(A2) For any symmetric $K\times K$ matrix $B$, $\mathrm{eigsp}(B)$ is defined as the minimum gap between its adjacent eigenvalues, i.e., $\mathrm{eigsp}(B)=\underset{1\leq i\leq K-1}{\mathrm{min}}|\lambda_{i+1}-\lambda_{i}|, \lambda_{1}>\lambda_{2}>\ldots> \lambda_{K}.$  Assume there exists a constant number $C>0$ such that when $n$ is sufficiently large, $\mathrm{eigsp}(\bar{D}P\bar{D})\geq C$.

(A3) There is a positive constant $h$ such that
$\underset{1\leq i,j \leq K}{\mathrm{max}}\{\|\theta^{(i)}\|/\|\theta^{(j)}\|\}\leq h.$

(A4) For sufficiently large $n$, we have
$\mathrm{log}(n)\theta^{2}_{\mathrm{max}}/\theta_{\mathrm{min}}\leq \|\theta\|^{3}_{3}.$

The following theorem provides a rough bound for $\|\hat{X}^{*}-X^{*}\|_{F}$, which is proved in Appendix.
\begin{thm}\label{mainPCC1}
	Under the DCSBM with parameters $\{n, P, \Theta, Z\}$ and the four assumptions (A1)-(A4) hold, with probability at least $1-o(n^{-3})$, we have
	\begin{align*}
	\|\hat{X}^{*}-X^{*}\|_{F}\leq 4\sqrt{n\mathrm{log}(n)\theta_{\mathrm{max}}\|\theta\|_{1}/\|\theta\|^{4}} +\sqrt{nC\mathrm{log}(n)\|\theta\|_{1}\|\theta\|_{3}^{3}/\|\theta\|^{6}},
	\end{align*}
	where $\theta_{\mathrm{max}}=\mathrm{max}_{i}\{\theta_{i}\}$ and $C$ is a constant.
\end{thm}
Now we give the bound of Hamming error for PCC. The Hamming error rate of PCC is defined as:
\begin{align*}
\mathrm{Hamm}_{n}(\hat{\ell},\ell)=\underset{\pi\in S_{K}}{\mathrm{min~}}H_{p}(\hat{\ell},\pi(\ell))/n,
\end{align*}
where $S_{K}=\{\pi:\pi \mathrm{~is~a~permutation~of~set~}\{1,2,\ldots, K\}\}$ \footnote{ Due to the fact that the clustering errors should not
	depend on how we tag each of the K communities, that's why we need to consider permutation to measure the clustering errors of PCC here.} and $H_{p}(\hat{\ell},\ell)$ is the expected number of mismatched labels and defined as below
\begin{align*}
H_{p}(\hat{\ell},\ell)=\sum_{i=1}^{n}P(\hat{\ell}(i)\neq \ell(i)).
\end{align*}
Then under some mild conditions, we build the bound for the Hamming error of PCC algorithm and show that the PCC method stably yields consistent community detection. We present the results in Theorem \ref{mainPCC2} and its proof in Appendix.
\begin{thm} \label{mainPCC2}
	Under the $DCSBM(n, P, \Theta, Z)$, assume (A1)-(A4) hold. Set \begin{align*}	err_{n}=4(4\sqrt{n\mathrm{log}(n)\theta_{\mathrm{max}}\|\theta\|_{1}/\|\theta\|^{4}}\\+ \sqrt{nC\mathrm{log}(n)\|\theta\|_{1}\|\theta\|_{3}^{3}/\|\theta\|^{6}})^{2}.
	\end{align*}
	Suppose as $n\rightarrow \infty$, we have
	\begin{align*}
	err_{n}/\mathrm{min~}\{n_{1}, n_{2}, \ldots, n_{K}\}\rightarrow 0,
	\end{align*}
	where $n_{k}$ is the size of the $k$-th community for $1\leq k\leq K$.
	For the estimated label vector $\hat{\ell}$ by PCC, with sufficiently large $n$, we have
	\begin{align*}
	\mathrm{Hamm}_{n}(\hat{\ell},\ell) \leq err_{n}/n.
	\end{align*}
\end{thm}
Theorem \ref{mainPCC2} is the main theoretical result for PCC, since it provides the theoretical bound for the Hamming error rate of PCC. By assumptions (A1)-(A4), we can find that $err_{n}/n$ goes to 0 as $n$ goes to infinity,  suggesting that the Hamming error rates for PCC decreases to 0 as the increasing of $n$, which guarantees that PCC yields stable consistent clustering.
\section{Normalized Principal Component Clustering (NPCC)}\label{sec:NPCC}
Our NPCC approach is designed based on the PCC and an asymmetric regularized Laplacian matrix. Therefore, firstly we introduce the asymmetric regularized Laplacian matrix.

\subsection{The Algorithm of NPCC}\label{algNPCC}
Define a diagonal matrix $D$ such that
\begin{align*}
  D_{ii}=\sum_{j=1}^{n}A_{ij},\qquad i=1,2,\ldots,n.
\end{align*}
The regularized Laplacian matrix $L_{\tau}$ can be defined by
\begin{align*}
  L_{\tau}=D_{\tau}^{-1/2}AD_{\tau}^{-1/2},
\end{align*}
where $D_{\tau}=D+\tau I$, $I$ is an identity matrix and $\tau\geq 0$ is a regularizer. Note that as suggested by \cite{RSC}, the default $\tau$ in the regularized Laplacian matrix is always set as the average node degree (i.e, $\tau=\sum_{i=1}^{n}D_{ii}/n$).

Now we give the general definition of the normalized matrix as following:
\begin{defin}\label{defN}
For any $s\times t$ matrix $Y=[Y_{1}, Y_{2}, \ldots, Y_{t}]$ where $Y_{i}$ denotes the $i$-th column of $Y$, define its column normalized matrix $N^{Y}$ such that the $i$-th column $N_{i}^{Y}$ is defined as
\begin{align*}
  N_{i}^{Y}=Y_{i}/\|Y_{i}\|_{2},\qquad 1\leq i\leq t.
\end{align*}
If $Y_{i}=0$, then define $N_{i}^{Y}$ as an abtrary nonzero vector with norm 1.
\end{defin}\label{defNormalized}
After defining column normalized matrix for any given matrix, now we can compute our normalized Laplacian matrix $N^{L_{\tau}}$ (the asymmetric regularized Laplacian matrix) via Definition \ref{defNormalized}. Note that since we only consider connected network (i.e., $A$ is a connected matrix without isolated nodes and disconnected parts), there are no zero columns in $L_{\tau}$, hence $N^{L_{\tau}}$ is well defined.

The spectral algorithm PCC is simple and without any parameters. In this section, we will show that via replacing the adjacency matrix $A$ by $N^{L_{\tau}}$ in PCC, we can design a method (i.e., NPCC) with significant improvements compared with PCC.  Therefore, NPCC is also designed based on the normalized Laplacian matrix which is related to the parameter $\tau$ where $\tau$ is originally introduced by \cite{RSCtau}. That's why we state that NPCC is a combination of PCC and RSC. 
Now we present the normalized principal component clustering algorithm as follows:
\begin{algorithm}
        \caption{ \textbf{Normalized principal component clustering(NPCC)}}
        \label{alg:NPCC}
        \begin{algorithmic}[1]
        \Require $A,K$ and regularizer $\tau$ (a good default for $\tau$ is the average node degree).
        \Ensure node labels $\check{\ell}$
        \State Compute $\check{X}=V_{N^{L_{\tau}}}E_{N^{L_{\tau}}}$, where $E_{N^{L_{\tau}}}$ is the $K\times K$ diagonal matrix containing the $K$ leading eigenvalues of $N^{L_{\tau}}$, and $V_{N^{L_{\tau}}}$ is the $n\times K$ matrix containing the corresponding  right eigenvectors with unit-norm.
        \State  Compute $\check{X}^{*}$, a normalized and regularized version of $\check{X}$, the rows of which are given by $\check{X}_{i}^{*}=\frac{\check{X}_{i}}{\|\check{X}_{i}\|_{2}}$.
        \State Obtain node labels $\check{\ell}$ by applying classical clustering method $k$-means to $\check{X}^{*}$ with assumption that there are $K$ clusters.
        \end{algorithmic}
\end{algorithm}

As suggested in \cite{RSCtau} and \cite{RSC}, a good default for $\tau$ is the average node degree. Later, we will study properties of NPCC under DCSBM and show that NPCC returns perfect partitions under the ideal case. The ideal case means that in Algorithm \ref{alg:NPCC}, the inputs are $(\Omega, K,\tau)$ instead of $(A, K, \tau)$. Compared with the PCC algorithm (Algorithm \ref{alg:PCC}), we can find that NPCC is designed based on an asymmetric matrix $N^{L_{\tau}}$, and NPCC also take advantage of the leading $K$ eigenvalues of $N^{L_{\tau}}$ to re-weight $\check{X}$.

Lemma \ref{realeigen} guarantees that $\check{X}$ is well defined.  Since $L_{\tau}$ is a symmetric matrix, we know that all eigenvalues and eigenvectors of $L_{\tau}$  are real numbers. Since $N^{L_{\tau}}=L_{\tau}U_{L_{\tau}}$ where $U_{L_{\tau}}$ is a diagonal matrix such that its $i$-th diagonal entry is $\frac{1}{\|L^{i}_{\tau}\|}$ for $1\leq i\leq n$ (where $L^{i}_{\tau}$ is the $i$-th column of $L_{\tau}$), by Lemma \ref{realeigen}, we can find that all eigenvalues and eigenvectors of $N^{L_{\tau}}$ are real numbers though $N^{L_{\tau}}$ is an asymmetric matrix, which guarantees that $\check{X}$ in Algorithm \ref{alg:NPCC} is well-defined (same conclusions hold for $\tilde{X}$ defined in next sub-section). The proof of Lemma \ref{realeigen} is given in Appendix.
\begin{lem}\label{realeigen}
For any $n\times n$ symmetric matrix $Y$ with rank $\mathrm{rank}(Y)$. Set $U_{Y}$ be the $n\times n$ diagonal matrix such that $U_{Y}(i,i)=1/\|Y_{i}\|, 1\leq i\leq n$. Assume that $U_{Y}$ is nonsingular,  then we have: $\mathrm{rank}(N^{Y})=\mathrm{rank}(Y)$, and all eigenvalues of $N^{Y}$ are real numbers. Furthermore, $Y$ and $N^{Y}$ share the same numbers of positive, negative, and zero eigenvalues.
\end{lem}
\subsection{Population analysis of the NPCC}\label{popNPCC}
This section shows that NPCC perfectly reconstructs the block partition under $DCSBM(n, P, \Theta, Z)$.

Let $\mathscr{D}$ be a $n\times n$ diagonal matrix such that $\mathscr{D}$ contains the expected node degrees, $\mathscr{D}_{ii}=\sum_{j}^{n}\Omega_{ij}$. Define the regularized version of $\mathscr{D}$ such that $\mathscr{D}_{\tau}=\mathscr{D}+\tau I$ where the regularizer $\tau$ is nonnegative. Then the population Laplacian matrix $\mathscr{L}$ and the regularized  population Laplacian matrix $\mathscr{L}_{\tau}$ in $\mathcal{R}^{n\times n}$ can be defined in the following way:
\begin{align*}
  \mathscr{L}=\mathscr{D}^{-0.5}\Omega \mathscr{D}^{-0.5},~~~\mathscr{L}_{\tau}=\mathscr{D}_{\tau}^{-0.5}\Omega \mathscr{D}_{\tau}^{-0.5}
\end{align*}
Define a $K\times n$ matrix $Q$ as $Q=PZ'\Theta$, and define a $K\times K$ diagonal matrix $D_{P}$ as $D_{P}(i,i)=\sum_{j=1}^{n}Q_{ij}, i=1,2,\ldots,K$. The explicit form of $\mathscr{L}_{\tau}$ is obtained via the following Lemma.
\begin{lem}\label{lemmaLtau}
(\emph{Explicit form for} $\mathscr{L}_{\tau}$) Under $DCSBM$ with parameters $\{n, P, \Theta, Z\}$, define $\theta_{i}^{\tau}$ as $\theta_{i}^{\tau}=\theta_{i}\frac{\mathscr{D}_{ii}}{\mathscr{D}_{ii}+\tau}$,
let $\Theta_{\tau}\in \mathcal{R}^{n\times n}$ be a diagonal matrix whose $ii$'th entry is $\theta_{i}^{\tau}$, for $i = 1, \dots, n$, and define $\tilde{P}=D_{P}^{-0.5}PD_{P}^{-0.5}$. Then $\mathscr{L}_{\tau}$ can be written as
\begin{align*}
  \mathscr{L}_{\tau}=\mathscr{D}_{\tau}^{-0.5}\Omega \mathscr{D}_{\tau}^{-0.5}=\Theta_{\tau}^{0.5}Z\tilde{P}Z'\Theta_{\tau}^{0.5}.
\end{align*}
\end{lem}
Lemma \ref{lemmaLtau} shows that $\mathscr{L}_{\tau}$ can be rewritten as a product of matrices related with the regularizer parameter $\tau$. This Lemma is equivalent to Lemma 3.2 in \cite{RSC}, so we omit the proof of it here. Meanwhile, note that under $DCSBM(n, P, \Theta, Z)$, we have $\Omega=\Theta ZPZ'\Theta$. By Lemma \ref{lemmaLtau}, $\mathscr{L}_{\tau}$ can be rewritten as a similar simple form as $\Omega$, which suggests that $\mathscr{L}_{\tau}$ may share similar properties as $\Omega$, and that's what we do on $\mathscr{L}_{\tau}$ next.

Then define a $n\times 1$ vector  $\tilde{\theta}$ such that its $i$-th entry is defined as $\tilde{\theta}_{i}=\sqrt{\theta_{i}^{\tau}}.$
Let $\tilde{\theta}^{(k)}$ be the $n\times 1$  vectors such that
$ \tilde{\theta}^{(k)}(i)=\tilde{\theta}(i)\mathrm{~or~}0,$  according to $g_{i}=k$ or not, for $1\leq k\leq K.$
Let $\tilde{D}$ be a $K\times K$ diagonal matrix of the \emph{overall degree intensities} $\tilde{D}(k,k)=\|\tilde{\theta}^{(k)}\|/\|\tilde{\theta}\|,$ for $ k  = 1, \dots, K.$ Now we define a $n\times K$ matrix $\tilde{\Gamma}$ such that
\begin{align*}
  \tilde{\Gamma}=[\frac{\tilde{\theta}^{(1)}}{\|\tilde{\theta}^{(1)}\|} ~ \frac{\tilde{\theta}^{(2)}}{\|\tilde{\theta}^{(2)}\|} ~\ldots~\frac{\tilde{\theta}^{(K)}}{\|\tilde{\theta}^{(K)}\|}].
\end{align*}
Then due to the special form of $\tilde{\Gamma}$, we have $\tilde{\Gamma}'\tilde{\Gamma}=I$, where $I$ is the $K\times K$ identity matrix. Now we can also write $\mathscr{L}_{\tau}$ as
\begin{align*}
\mathscr{L}_{\tau}=\|\tilde{\theta}\|_{2}^{2}\tilde{\Gamma}\tilde{D}\tilde{P}\tilde{D}\tilde{\Gamma}'.
\end{align*}
The above special form for $\mathscr{L}_{\tau}$ plays an important role for the proof of Lemma \ref{mainNPCC1} given below. After we have $\tilde{\Gamma}$, we can define a $K\times K$ matrix $\tilde{F}$ as $\tilde{F}=\tilde{\Gamma}'N^{\mathscr{L}_{\tau}}\tilde{\Gamma}$. According to the special form of $\tilde{\Gamma}$ and $\tilde{F}$, we have Lemma \ref{mainNPCC1} which shows that $N^{\mathscr{L}_{\tau}}$ shares the same nonzero eigenvalues of $\tilde{F}$, and we can use the eigenvectors of $\tilde{F}$ to give the expressions of the leading $K$ eigenvectors of $N^{\mathscr{L}_{\tau}}$.
\begin{lem}\label{mainNPCC1}
Under $DCSBM(n, P, \Theta, Z)$, let $\tilde{\lambda}_{1}, \tilde{\lambda}_{2}, \ldots, \tilde{\lambda}_{K}$ be the $K$ nonzero  eigenvalues of $\tilde{F}$ (arranged in the descending order of the magnitudes), and let $\tilde{a}_{1}, \tilde{a}_{2}, \ldots, \tilde{a}_{K}$ be the associated (unit-norm) right eigenvectors. Then we have:

 1) $\tilde{\Gamma}\tilde{F}=N^{\mathscr{L}_{\tau}}\tilde{\Gamma}$,  where $N^{\mathscr{L}_{\tau}}$ is the column normalized matrix of $\mathscr{L}_{\tau}$;

 2) the $K$ nonzero eigenvalues of $N^{\mathscr{L}_{\tau}}$ are  $\tilde{\lambda}_{1}, \tilde{\lambda}_{2}, \ldots, \tilde{\lambda}_{K}$, with the associated unit-norm right eigenvectors being
\begin{align*}
  \tilde{\eta}_{k}=\sum_{i=1}^{K}[\tilde{a}_{k}(i)/\|\tilde{\theta}^{(i)}\|_{2}]\tilde{\theta}^{(i)},\qquad k=1,2,\ldots, K.
\end{align*}
\end{lem}\label{lemmaNLtau}

The proof of Lemma \ref{mainNPCC1} can be found in the Appendix.  Set $\tilde{X}=V_{N^{\mathscr{L}_{\tau}}}E_{N^{\mathscr{L}_{\tau}}}$, where $E_{N^{\mathscr{L}_{\tau}}}$ is the $K\times K$ diagonal matrix containing the $K$ nonzero real eigenvalues of $N^{\mathscr{L}_{\tau}}$, and $V_{N^{\mathscr{L}_{\tau}}}$ is the $n\times K$ matrix containing the corresponding right eigenvectors with unit-norm. We'd emphasize that Lemma \ref{mainNPCC1} is significant due to the fact that we can obtain the expressions of each entries of $\tilde{X}$ via Lemma \ref{mainNPCC1} under $DCSBM(n, P, \Theta, Z)$. Then set $\tilde{X}^{*}$ from $\tilde{X}$ by normalizing each of $\tilde{X}$'s rows to have unit length.

After we have defined $\tilde{X}^{*}$, for the population analysis of NPCC, we write down the Ideal NPCC algorithm. Input: $\Omega$. Output: $\tilde{\ell}$.
\begin{itemize}
  \item Obtain $\tilde{X}=V_{N^{\mathscr{L}_{\tau}}}E_{N^{\mathscr{L}_{\tau}}}$.
  \item Obtain $\tilde{X}^{*}$
  \item Applying k-means to $\tilde{X}^{*}$ assuming there are $K$
  communities to obtain $\tilde{\ell}$.
\end{itemize}
Therefore, the population analysis of NPCC method can be defined as:
following the Ideal NPCC algorithm
and answer the question that whether the Ideal NPCC algorithm returns perfect clustering under $DCSBM$. The answer is YES and it is guaranteed by the following lemma:
\begin{lem}\label{mainNPCC2}
Under $DCSBM(n, P, \Theta, Z)$, $\tilde{X}^{*}$ has exactly $K$ distinct rows, and if $g_{i}=g_{j}$ for any node $i\neq j$, then the $i$-th  row of $\tilde{X}^{*}$ equals to the $j$-th row of $\tilde{X}^{*}$.
\end{lem}
This lemma guarantees that running k-means on the rows of $\tilde{X}^{*}$ will return perfect clustering. The proof of Lemma \ref{mainNPCC2} can be found in the Appendix.

\begin{figure}[H] \includegraphics[width=\textwidth]{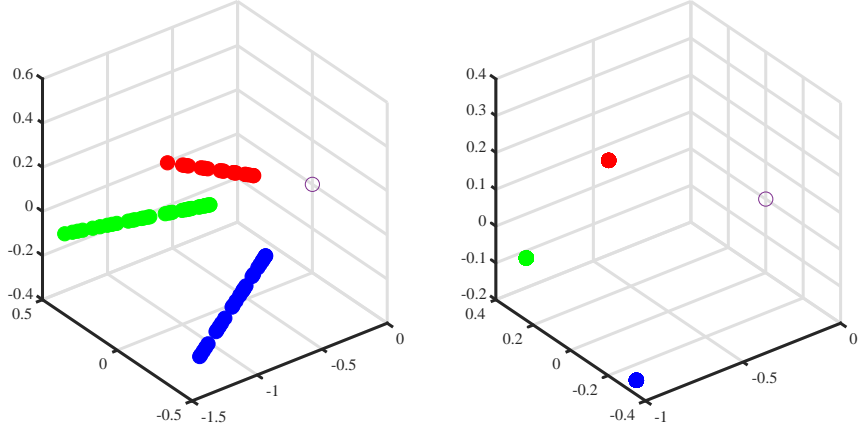}
	\caption{In this synthetic data example., $\tilde{X}$ comes from the DCSBM with parameters setting as: $(n,K)=(90,3)$, the mixing matrix $P$ has diagonal entries as 0,6 and off-diagonal entries as 0.3, nodes belong to three communities with equal probability and $\theta(i)=0.3+0.7\times (i/n)^{2}$ for $1\leq i\leq n$. Each point corresponds to one row of the matrix $\tilde{X}$(in the left panel) or $\tilde{X}^{*}$ (in the right panel). The different colors indicate three different communities. The
hollow circle denotes the origin point. Without normalization (left panel), nodes belonging to the same community share the \emph{\textbf{same direction}} in the spatial coordinate system. After normalization (right panel), nodes belonging to same community share the \emph{\textbf{same position}}
in the spatial  coordinate system.}\label{ExpectationNPCC}
\end{figure}

An illustration about the geometry property of $\tilde{X}$ and $\tilde{X}^{*}$ under the case that there are three communities in a simulated network is given by Figure \ref{ExpectationNPCC}. From this figure, we can find that $\tilde{X}^{*}$ has exactly $K$ distinct rows and applying k-means on $\tilde{X}^{*}$ return true labels for all nodes. For PCC method, we have similar figures for $X$ and $X^{*}$ under the same setting as Figure \ref{ExpectationNPCC}, we omit it here to save space.
\subsection{Spectral analysis of $N^{L_{\tau}}$}\label{spectralNPCC}
This section gives the expression of  the leading $K$ right eigenvectors of $N^{L_{\tau}}$. Define the $n\times n$ diagonal matrix $U_{\mathscr{L}_{\tau}}$ as
\begin{align*}
 U_{\mathscr{L}_{\tau}}(i,i)=\frac{1}{\|\mathscr{L}_{\tau}^{i}\|_{2}},\qquad i=1,2,\ldots, n,
\end{align*}
where $\|\mathscr{L}_{\tau}^{i}\|_{2}$ means the $\ell_{2}$ norm of the $i$-th column of $\mathscr{L}_{\tau}$. Then we can find that $N^{\mathscr{L}_{\tau}}=\mathscr{L}_{\tau} U_{\mathscr{L}_{\tau}}$.

Let $\check{\lambda}_{1},\check{\lambda}_{2},\ldots, \check{\lambda}_{K}$ be the $K$ leading eigenvalues of $N^{L_{\tau}}$. For $1\leq k\leq K$, if $\check{\lambda}_{k}$ is not an eigenvalue of  $N^{L_{\tau}}-N^{\mathscr{L}_{\tau}}$, let $B^{(k)}$ be a $K\times K$ matrix
\begin{align*}
  &B^{(k)}(i,j)=\\
  &(\|\tilde{\theta}^{(i)}\|\|\tilde{\theta}^{(j)}\|)^{-1}
  (\tilde{\theta}^{(i)})'U_{\mathscr{L}_{\tau}}[I_{n}-(N^{L_{\tau}}-
  N^{\mathscr{L}_{\tau}})/\check{\lambda}_{k}]^{-1}\tilde{\theta}^{(j)},
\end{align*}
where $1\leq i,j\leq K$. If $\check{\lambda}_{k}$ is an eigenvalue of $N^{L_{\tau}}-N^{\mathscr{L}_{\tau}}$, let $B^{(k)}$ be a $K\times K$ matrix of 0.
\begin{lem}\label{mainNPCC3}
Under $DCSBM(n, P, \Theta, Z)$, let $\{\check{\lambda}_{k}\}_{k=1}^{K}$ be the eigenvalues of $N^{L_{\tau}}$ with the largest magnitudes. Then the associated right eigenvector is given by
\begin{align*}
  \check{\eta}_{k}=\sum_{l=1}^{K}(e_{k}(l)/\|\tilde{\theta}^{(l)}\|)[I_{n}-(N^{L_{\tau}}-N^{\mathscr{L}_{\tau}})/\check{\lambda}_{k}]^{-1}\tilde{\theta}^{(l)},
\end{align*}
where $e_{k}$ is an (unit-norm) eigenvector of $\tilde{D}\tilde{P}\tilde{D}B^{(k)}$, and $\check{\lambda}_{k}$ is the unique eigenvalue of $\tilde{D}\tilde{P}\tilde{D}B^{(k)}$ that is associated with $e_{k}$.
\end{lem}
The full theoretical framework of NPCC follows a similar process as PCC. Therefore, the ideal full theoretic analysis of NPCC should include the following steps: in the first step we need to bound $\|\check{X}-\tilde{X}\|_{F}$ for NPCC, then we can bound $\|\check{X}^{*}-\tilde{X}^{*}\|_{F}$, finally the bound of $\|\breve{X}^{*}-\tilde{X}^{*}\|_{F}$ can give us the bound of the Hamming error rate of NPCC. Unfortunately, due to a theoretic gap of the NPCC method, we can't give the full theoretic analysis of NPCC at present. Now we present the theoretic gap briefly. We move to the first step, find bound of $\|\check{X}-\tilde{X}\|_{F}$ for NPCC: in order to bound
$\|\check{X}-\tilde{X}\|_{F}$, we need to bound
$\|E_{N^{L_{\tau}}}-E_{N^{\mathscr{L}_{\tau}}}\|_{F}$. However, since Weyl's inequality \citep{Weyl} never holds for the two asymmetric matrices $N^{L_{\tau}}$ and $N^{\mathscr{L}_{\tau}}$, therefore we can't find the bound of $\|E_{N^{L_{\tau}}}-E_{N^{\mathscr{L}_{\tau}}}\|_{F}$ , i.e., we can't find the bound of $|\check{\lambda}_{k}-\tilde{\lambda}_{k}|$ for $1\leq k\leq K$ due to the fact that Weyl's inequality never holds on asymmetric matrices, and other available mathematical techniques also don't work. This theoretic gap prevents us from giving a full theoretic analysis of the NPCC method (not to mention that give the bound of Hamming error rate for NPCC). The detail about the statement on the theoretical gap for NPCC is given in the appendix \ref{TheoreticalGap}. For readers' reference, the expressions of the leading $K$ eigenvectors $\{\hat{\eta}_{k}\}_{k=1}^{K}$ of $A$ is given in Lemma 2.4 of \cite{SCORE}. Though we can not give full theoretical analysis for NPCC at present, we apply Lemma \ref{mainNPCC3} as future work's reference for completing the theoretic framework for NPCC.
\section{Simulation and Analysis of Real World Datasets}\label{dataanalysis}
In this section, we provide numerical results on both simulated and real-world datasets to investigate the performance of PCC and NPCC via comparing it with SCORE, RSC and OCCAM. We use the Hamming error rate to measure the clustering error rate for each method. The error rate is defined as 
\begin{align*}
  \mathrm{min}_{\{\pi: \mathrm{permutation~over~}\{1,2,\ldots, K\}\}}\frac{1}{n}\sum_{i=1}^{n}1\{\pi(\hat{\ell}(i))\neq \ell(i)\},
\end{align*}
where $\ell(i)$ and $\hat{\ell}(i)$ are the true and estimated labels of node $i$.
\subsection{Synthetic data experiments}
\subsubsection{Experiment 1}\label{Experiment1}
This experiment investigates how these approaches perform under SBM when $n$ (size of the simulated network) increases. Set $n\in \{100, 200, 300, 400, 500, 600\}$. For each fixed $n$, we record the mean of the error rates of 100 repetitions. This experiment contains two sub-experiments: Experiment 1(a) and 1(b).

\emph{Experiment 1(a)}. We test how these methods perform under SBM when $K=2$ in this sub-experiment. Generate $\ell$ by setting each node belonging to one of the clusters with equal probability (i.e., $\ell_{i}-1 \overset{\mathrm{i.i.d.}}{\sim}\mathrm{Bernoulli}(1/2)$). And the mixing matrix $P_{1(a)}$ is set as
 \[
P_{1(a)}
=
\begin{bmatrix}
    0.9&0.3\\
    0.3&0.8\\
\end{bmatrix}.
\]
Generate $\theta$ as $\theta(i)=0.2$ for $g_{i}=1$, $\theta(i)=0.6$ for $g_{i}=2$.

\begin{figure}[H]
	\includegraphics[width=\textwidth]{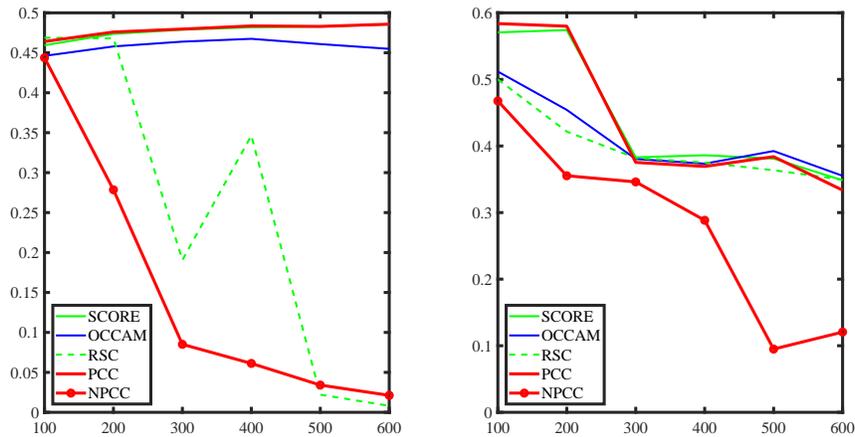}
	\caption{Comparison of Hamming error rates: left panel (Experiment 1(a)), right panel (Experiment 1(b). x-axis: $n$. y-axis: error rates.}\label{Ex1}
\end{figure}

Numerical results of Experiment 1(a) are showed by the left panel of Figure \ref{Ex1}, from which we can find that 1) NPCC outperforms PCC, SCORE,RSC and OCCAM in this sub-experiment. Meanwhile, PCC, SCORE and OCCAM fail to detect communities in this experiment since their error rates are always larger than 0.4. And since NPCC is designed based on PCC, from the left panel of Figure \ref{Ex1}, we can say that NPCC provides a significant improvement compared with PCC for Experiment 1(a). 2) As the size of network increases, NPCC and RSC have increasing performances with smaller error rates while PCC, SCORE and OCCAM fail to detect.

\emph{Experiment 1(b)}. We investigate how these methods perform with SBM when $K=3$ in this sub-experiment. Generate $\ell$  by setting each node belonging to one of the clusters with equal probability. And the mixing matrix $P_{1(b)}$ is set as
 \[
P_{1(b)}
=
\begin{bmatrix}
    0.9&0.3&0.3\\
    0.3&0.8&0.3\\
    0.3&0.3&0.7
\end{bmatrix}.
\]
Generate $\theta$ as $\theta(i)=0.2$ for $g_{i}=1$, $\theta(i)=0.4$ for $g_{i}=2$, and $\theta(i)=0.8$ for $g_{i}=3$.  

Numerical results of Experiment 1(b) are showed by the right panel of Figure \ref{Ex1}, from which we can find that 1) NPCC performs outperforms the other four approaches and PCC have competitive performances compared with SCORE. Meanwhile, NPCC always performs better than PCC. 2) With the increasing of sample size $n$, error rates for all approaches decreases.
\subsubsection{Experiment 2}
In this experiment, we study how the ratios of
 $\theta$, the ratios of the size between different communities and the ratios of diagonal and off-diagonal entries of the mixing matrix impact behaviors of our PCC and NPCC approaches when $n=400$ and $K=2$.

\textit{Experiment 2(a)}. We study  the influence of ratios of $\theta$ when $K=2$ under SBM in this sub-experiment. Set the proportion $a_{0}$ in $\{1, 2, \ldots, 8\}$. We generate $\ell$  by setting each node belonging to one of the clusters with equal probability. And the mixing matrix is $P_{2(a)}$.
 \[\renewcommand{\arraystretch}{0.75}
P_{2(a)}
=
\begin{bmatrix}
    0.9&0.4\\
    0.4&0.8\\
\end{bmatrix}.
\]
Generate $\theta$ as $\theta(i)=1$ if $g_{i}=1$ ,$\theta(i)=1/a_{0}$ for $g_{i}=2$. Note that for each fixed $a_{0}$, $\theta(i)$ is a fixed number for all nodes in the same community and hence this is a SBM case. Meanwhile, for each fixed $a_{0}$, we record the mean of clustering error rate of 100 sampled networks.

\begin{figure}[H]
	\includegraphics[width=\textwidth,height=6cm]{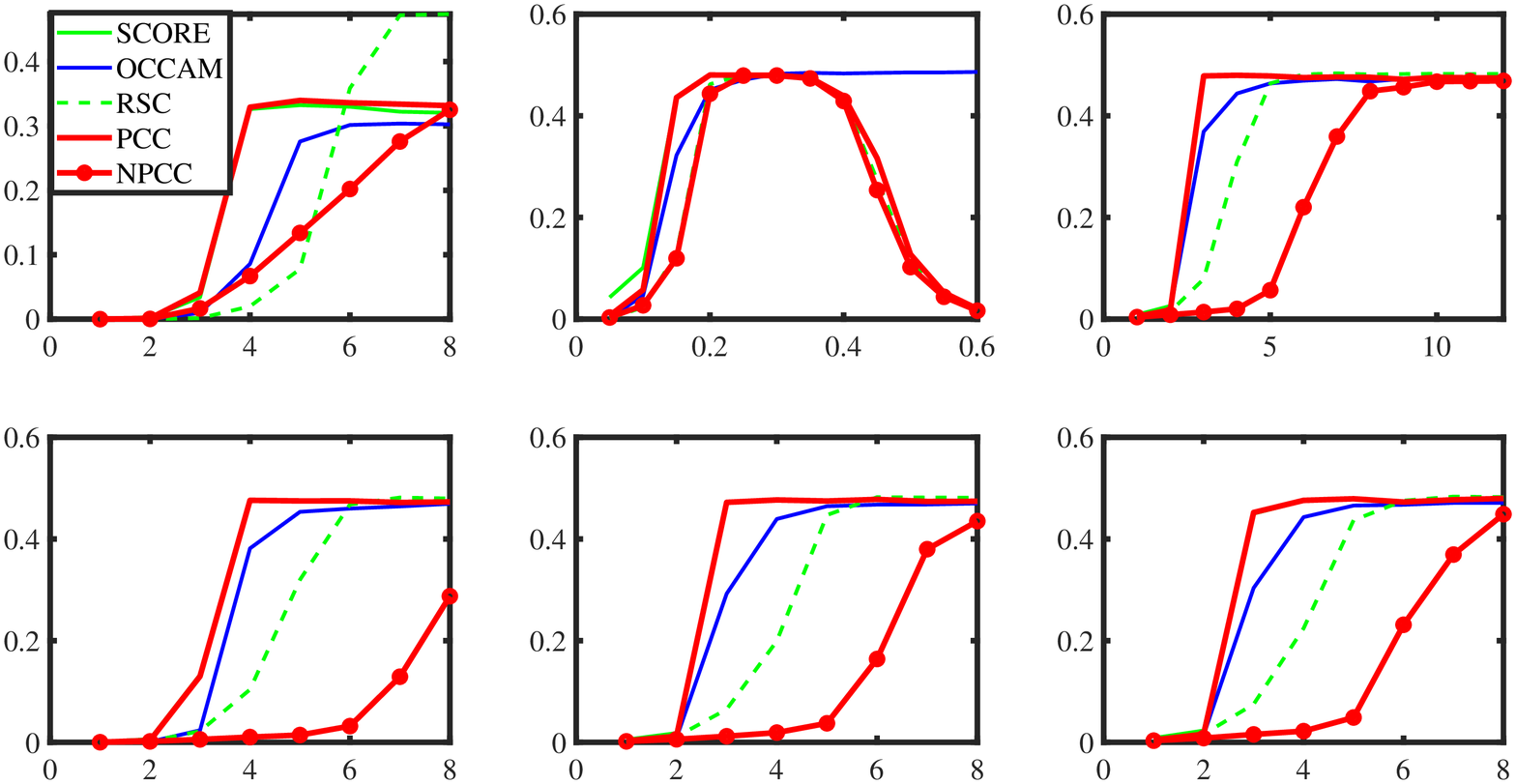}
	\caption{Comparison of Hamming error rates: top left panel (Experiment 2(a), x-axis: $a_{0}$), top middle panel  (Experiment 2(b), x-axis: $b_{0}$), top right panel (Experiment 2(c), x-axis: $c_{0}$). bottom left panel (Experiment 2(d), x-axis: $c_{0}$), bottom middle panel  (Experiment 2(e), x-axis: $c_{0}$), bottom right panel (Experiment 2(f), x-axis: $c_{0}$). For all six panels, y-axis: error rates.}\label{Ex2}
\end{figure}

The numerical results of Experiment 2(a) are showed by the top left panel of  Figure \ref{Ex2} from which we can find that 1) as $a_{0}$ increases ( which leads to a more heterogeneous case, and hence a case that is more difficult to be detected), the clustering error rates for all approaches increase. 2) When the ratios between two clusters are small (i.e., $1\leq a_{0}\leq 3$), all methods have satisfactory performances and behave similarly; while, when $a_{0}\geq 4$, NPCC and RSC have better performances compared with PCC, SCORE and OCCAM. Meanwhile, when $a_{0}$ is in the interval $[3,5]$, the RSC algorithm performs slightly better than NPCC. However, as $a_{0}$ continues increasing (i.e., $a_{0}\geq 5$), NPCC outperforms all the other three approaches. Especially, when $a_{0}$ is bigger than 6, the RSC method even performs poorer than PCC, SCORE and OCCAM while at the same time NPCC has the best performances. Finally, PCC shares similar performances as SCORE in this sub-experiment.

\textit{Experiment 2(b)}. We study how the proportion between the diagonals and off-diagonals of $P$ affects the performance of these methods under SBM in this sub-experiment. Set the proportion $b_{0}$  in $\{1/20, 2/20, 3/20,$ $\ldots, 9/20, 12/20\}$. We generate $\ell$ such that for $g_{i}=1$ for $1\leq i\leq n/2$; for $(\frac{n}{2}+1)\leq i\leq n,~g_{i}=2$.  Let $\theta(i)=0.4$ if $g_{i}=1$ and $\theta(i)=0.6$ otherwise. The mixing matrix $P_{2(b)}$ is
 \[\renewcommand{\arraystretch}{0.75}
P_{2(b)}
=
\begin{bmatrix}
   0.3& b_{0} \\
    b_{0} & 0.3\\
\end{bmatrix}.
\]
For each fixed $b_{0}$, we record the average for the clustering error rate of 100 simulated networks.

The numerical results of Experiment 2(b) are showed by the top middle panel of Figure \ref{Ex2}, which suggests us that 1) as $P$'s off-diagonal entry becomes closer to its diagonal entry, it is more challenge to detect community information for all methods since the clustering error rates for the four approaches increase as $b_{0}$ increases. 2) When $b_{0}\leq 0.3$, PCC and SCORE have the poorest performances compared with NPCC, RSC, and OCCAM. When $b_{0}\geq 0.3$, NPCC performs slightly better than PCC, SCORE and RSC while OCCAM fail to detect communities, which suggests that our two approaches NPCC and PCC as well as the two comparison procedures SCORE and RSC can detect both associative and dis-associative networks \footnote{Where dis-associative networks is defined as networks whose off diagonal entries of the mixing matrix $P$ are larger than the diagonal entries, suggesting that there are more edges in cross communities than within the same communities in a dis-associative network.}  while OCCAM can not detect dis-associative networks.

\textit{Experiment 2(c)}. We study how the proportion between the size of clusters influences the performance of these methods under SBM in this sub-experiment. We set the proportion $c_{0}$ in $\{1,2,\ldots,12\}$. Set $n_{1}= \mathrm{round}(\frac{n}{c_{0}+1})$
as the number of nodes in cluster 1 where $\mathrm{round}(x)$ denotes the nearest integer for any real number $x$. We generate $\ell$ such that for $i=1:n_{1}, ~g_{i}=1$;
for $i=(n_{1}+1):n, ~g_{i}=2$. Note that $c_{0}$ is the ratio \footnote{Number of nodes in cluster 2 is $n-\mathrm{round}(\frac{n}{c_{0}+1})\approx n-\frac{n}{c_{0}+1}=c_{0}\frac{n}{c_{0}+1}$, therefore
number of nodes in cluster 2 is $c_{0}$ times of that in cluster 1.} of the sizes of
cluster 2 and cluster 1. And the mixing matrix $P_{2(c)}$ is set as follows:
  \[\renewcommand{\arraystretch}{0.75}
P_{2(c)}
=
\begin{bmatrix}
    0.9&0.4\\
    0.4&0.8
\end{bmatrix}.
\]
Let $\theta$ be $\theta(i)=0.4$ if $g_{i}=1$ and $\theta(i)=0.6$ otherwise. Meanwhile, for each fixed $c_{0}$, we record the average for the clustering error rate of 100 simulated networks.

Experiment 2(d): we set $c_{0}\in\{1,2,\ldots, 8\}$, and $\theta_{i}=0.4+0.5(i/n)$ for $1\leq i\leq n$. All other parameters are same as Experiment 2(c). This experiment is designed under DCSBM.

Experiment 2(e): set $\theta_{i}=0.4+0.5(i/n)^{2}$ for $1\leq i\leq n$, keep all other parameters the same as Experiment 2(d).

Experiment 2(f): set $\theta_{i}=0.4+0.5(i/n)^{3}$ for $1\leq i\leq n$, keep all other parameters the same as Experiment 2(d).

The numerical results of Experiments 2(c), 2(d), 2(e) and 2(f) are shown by the last four panels of Figure \ref{Ex2} from which we see that 1) when $n$ is fixed at 400, increasing the size of cluster 2 (therefore the size of cluster 1 is decreasing), it becomes more challenge to detect nodes from cluster hence all approaches behave poorer with increasing clustering error rates. 2) NPCC outperforms all the other methods in this sub-experiment whether under SBM or DCSBM. Meanwhile, RSC performs better than PCC, SCORE and OCCAM. PCC, SCORE and OCCAM have similar behaviors in this sub-experiment. Numerical results of this four sub-experiments suggest that NPCC returns more accurate detection information when dealing with networks with various size of clusters.

\subsubsection{Experiment 3}
In this experiment, we compare the computing time of PCC, NPCC with SCORE, OCCAM, RSC and CMM. For comparison of computing time, we'd use large networks. The detail of parameters under $DCSBM(n, P, \Theta, Z)$ are:
generate $\ell$ by setting each node belonging to one of the clusters with equal probability. Set the mixing matrix $P_{3}$ as
 \[
P_{3}
=
\begin{bmatrix}
    1&0.5&0.5&0.5\\
    0.5&1&0.5&0.5\\
    0.5&0.5&1&0.5\\
    0.5&0.5&0.5&1
\end{bmatrix}.
\]
Generate $\theta$ as $\theta(i)=0.2$ for $g_{i}=1$, $\theta(i)=0.4$ for $g_{i}=2$, $\theta(i)=0.6$ for $g_{i}=3$, $\theta(i)=0.8$ for $g_{i}=4$. Set $n\in \{500, 1000, 2000, 3000, 4000\}$. And for each fixed $n$, we record the error rates and computing time of the average of 10 independent repetitions.
\begin{table}[h!]
	\centering
	\caption{Comparison of error rates and computation time on simulated data for Experiment 3.}
	\label{tab:table1}
	\resizebox{\columnwidth}{!}{
	\begin{tabular}{cccccccccc}
		\toprule
		$n$ &PCC&NPCC&SCORE&OCCAM&RSC&CMM\\
		\midrule
        500&0.3244(0.47s)&0.3080(0.51s)&0.3266(0.42s)&0.3178(0.43s)&0.3134(0.54s)&0.1020(22s)\\
		1000&0.1693(0.67s)&0.2595(1.02s)&0.2791(0.71s)&0.2824(0.85s)&0.2743(1.12s)&0.0833(115s)\\
		2000&0.0429(1.13s)&0.0215(3.04s)&0.2824(1.55s)&0.5315(3.06s)&0.2666(4.06s)&0.0462(822s)\\
        3000&0.0176(2.09s)&0.0053(7.88s)&0.2790(2.88s)&0.0235(7.64s)&0.5619(8.29s)&0.0196(2560s)\\
		4000&0.0077(3.64s)&0.0013(18.29s)&0.2742(4.95s)&0.6218(17.34s)&0.0035(18.68s)&0.0082(6799s)\\
		\bottomrule
	\end{tabular}}
\end{table}

The numerical results of Experiment 3 are recorded in Table \ref{tab:table1} from which we can find that both PCC and NPCC provide satisfactory performances in error rates and computing time compared with SCORE, OCCAM, RSC and CMM. Though the numerical performance of CMM is satisfactory, CMM is pretty time-consuming due to its optimization procedure.
\subsection{Application to real-world datasets}
We use the four real-world datasets to investigate the performances of PCC and NPCC via comparison with SCORE, OCCAM, RSC and CMM.  Firstly, we briefly introduce these four networks.
\begin{itemize}
	\item \textbf{Karate club}: this network consists of 34 nodes where each
	node denotes a member in the karate club \citep{karate}. As there is a conflict in the club, the network divides into two communities: Mr. Hi’s group and John’s group. \cite{karate} records all labels for each member and we use them as the true labels.
	\item \textbf{Political blog network dataset}: this network consists of political blogs during the
	2004 US presidential election \citep{Polblogs1}. Each blog belongs to one of the two parties liberal or
	conservative. As suggested by \cite{DCSBM}, we only consider the largest connected component with 1222  nodes and ignore the edge direction for community detection.
	\item \textbf{Simmons College network}: this network contains one largest connected
	component with 1137 nodes. It is observed in \cite{traud2011comparing, traud2012social} that the community structure of the Simmons College network exhibits a strong correlation with the graduation year-students since students in the same year are more likely to be friends.
	\item \textbf{Caltech network}: this network has one largest connected component with 590 nodes. The community structure is highly correlated with which of the 8 dorms a user is from, as observed in \cite{traud2011comparing, traud2012social}.
\end{itemize}
Table \ref{tab:table2} presents some basic information of these four data sets \footnote{$\bar{d}$ denotes the average degree (i.e., $\bar{d}=\sum_{i,j=1}^{n}A_{ij}/n$)}.
\begin{table}[h!]
	\footnotesize
	\centering
	\caption{4 real world network datasets}
	\label{tab:table2}
	\begin{tabular}{cccccccccc}
		\toprule
		Dataset &\# Nodes &K&\#Edges &$\bar{d}$\\
		\midrule
		Karate&34&2&159&4.59\\
		Weblogs&1222&2&16714&27.35\\
		Simmons&1137&4&24257&42.67\\
		Caltech&590&8&12822&43.36\\
		\bottomrule
	\end{tabular}
\end{table}
From \cite{SCORE+}, we know that Simmons anf Caltech are two weak signal networks whose $(K+1)$-th leading eigenvalue is close to the $K$-th one of $A$ or its variants, while Karate and Weblogs are two strong signal networks. Suggested by \cite{SCORE+}, when dealing with weak signal networks, since the $(K+1)$-th leading eigenvectors may also contain information about nodes labels, therefore we conduct $(K+1)$ leading eigenvectors and eigenvalues of NPCC and PCC for clustering. Based on the above analysis, we provide two algorithms PCC+ and NPCC+ as the refinements of PCC and NPCC, respectively. The detail of the two refinements algorithms PCC+ and NPCC+ are given in Algorithm \ref{alg:PCC+} and Algorithm \ref{alg:NPCC+}, respectively.
\begin{algorithm}
        \caption{\textbf{PCC+}}
        \label{alg:PCC+}
        \begin{algorithmic}[1]
        \Require $A, K$ and threshold $t>0$.
        \Ensure $\hat{\ell}$.
        \State Set $M=K+1$, if $1-|\hat{\lambda}_{K+1}/\hat{\lambda}_{K}|<t$, otherwise set it as $K$
        \State Compute $\hat{X}=V_{A}E_{A}$, where $E_{A}$ is an $M\times M$ diagonal matrix that contains the leading $M$ eigenvalues of $A$ a, and $V_{A}$ is an $n\times M$ matrix contains the respective unit norm eigenvector of $A$ as column.
        \State  Compute $\hat{X}^{*}$ by normalizing each row of $\hat{X}$ to have unit length.
        \State  Apply k-means to $\hat{X}^{*}$ assuming there are $K$ clusters to obtain $\hat{\ell}$.
        \end{algorithmic}
\end{algorithm}

\begin{algorithm}
        \caption{\textbf{NPCC+}}
        \label{alg:NPCC+}
        \begin{algorithmic}[1]
        \Require $A, K$, regularizer $\tau$ and threshold $t>0$.
        \Ensure $\hat{\ell}$.
        \State Set $M=K+1$, if $1-|\check{\lambda}_{K+1}/\check{\lambda}_{K}|<t$, otherwise set it as $K$
        \State Compute $\check{X}=V_{N^{L_{\tau}}}E_{N^{L_{\tau}}}$, where $E_{N^{L_{\tau}}}$ is the $M\times M$ diagonal matrix containing the $M$ leading eigenvalues of $N^{L_{\tau}}$, and $V_{N^{L_{\tau}}}$ is the $n\times M$ matrix containing the corresponding  right eigenvectors with unit-norm.
        \State  Compute $\check{X}^{*}$ by normalizing each row of $\check{X}$ to have unit length.
        \State  Apply k-means to $\check{X}^{*}$ assuming there are $K$ clusters to obtain $\check{\ell}$.
        \end{algorithmic}
\end{algorithm}

\begin{table}[h!]
\footnotesize
\centering
\caption{Error rates on the four datasets}
 \label{tab:errorreal}
\begin{tabular}{cccccccccc}
 \toprule
 \textbf{ Methods} &Karate&Weblogs&Simmons&Caltech\\
 \midrule
SCORE&1/34&\textbf{58/1222}&268/1137&183/590\\
OCCAM&\textbf{0/34}&65/1222&266/1137&189/590\\
RSC&\textbf{0/34}&64/1222&244/1137&170/590\\
CMM&\textbf{0/34}&62/1222&137/1137&124/590\\
\hline
\textbf{PCC}&\textbf{0/34}&60/1222&243/1137&166/590\\
\textbf{PCC+}&\textbf{0/34}&60/1222&156/1137&97/590\\
\textbf{NPCC}&\textbf{0/34}&62/1222&225/1137&155/590\\
\textbf{NPCC+}&\textbf{0/34}&62/1222&\textbf{121/1137}&\textbf{96/590}\\
\bottomrule
\end{tabular}
\end{table}
The analysis results of these approaches are summarized in Table \ref{tab:errorreal} for the four networks. From Table \ref{tab:errorreal}, we can find that 1) for Karate club and Political blog networks, all methods have good performance and behave similarly.  Since we know that SCORE has the best-known number of error 58 for Political blogs network in the literature \cite{SCORE}, we can find that our NPCC and PCC have comparable performance to the state-of-the-art; 2) for Simmons and Caltech networks, PCC and NPCC have better performances than SCORE, OCCAM and RSC. Though CMM outperforms NPCC and PCC on Simmons and Caltech, its computing times for this two data sets are much larger than that of NPCC and PCC. By considering one more eigenvectors for clustering, we find that our PCC+ has obvious improvements  than PCC, our NPCC+ dramatically outperforms all other methods in this paper on the two weak signal networks Simmons and Caltech, which suggests that NPCC+ can detect weak signal networks efficiently.

By observing PCC+ and NPCC+, we find  that more than applying $(K+1)$ eigenvectors for clustering, our PCC and NPCC approaches has a nice property: in the 2nd steps in Algorithm \ref{alg:PCC} and Algorithm \ref{alg:NPCC}, $\hat{X}$ and $\check{X}$ can be computed with more than $K$ eigenvectors and eigenvalues for clustering, which suggests that if we add other more  vectors to compute $\hat{X}$ and $\check{X}$, there is little influence on the clustering results for both PCC and NPCC.  This property can be shown by the following two algorithms (we call them as $\mathrm{PCC}^{*}$ and $\mathrm{NPCC}^{*}$):
\begin{algorithm}
        \caption{$\mathrm{PCC}^{*}$}
        \label{alg:PCCstar}
        \begin{algorithmic}[1]
        \Require $A,K$
        \Ensure $\hat{\ell}$
        \State Compute $\hat{X}=V_{A}E_{A}$ where $E_{A}$ contains the leading $M_{k}$ eigenvalues of $A$ and $V_{A}$ contains the respective unit norm eigenvectors. Note that $M_{k}\geq K$.
        \State  Compute $\hat{X}^{*}$ by normalizing each of $\hat{X}$' rows to have unit length.
         \State Apply k-means to $\hat{X}^{*}$ assuming there are $K$ clusters to obtain $\hat{\ell}$.
        \end{algorithmic}
\end{algorithm}
\begin{algorithm}
        \caption{$\mathrm{NPCC}^{*}$}
        \label{alg:NPCCstar}
        \begin{algorithmic}[1]
        \Require $A,K$
        \Ensure $\hat{\ell}$
        \State Compute $\check{X}=V_{N^{L_{\tau}}}E_{N^{L_{\tau}}}$ where $E_{N^{L_{\tau}}}$ contains the leading $M_{k}$ eigenvalues of $N^{L_{\tau}}$ and $V_{N^{L_{\tau}}}$ contains the respective right unit norm eigenvectors. Note that $M_{k}\geq K$.
        \State  Compute $\check{X}^{*}$ by normalizing each of $\check{X}$' rows to have unit length.
         \State Apply k-means to $\check{X}^{*}$ assuming there are $K$ clusters to obtain $\check{\ell}$.
        \end{algorithmic}
\end{algorithm}

\begin{figure}[H]
	\includegraphics[width=\textwidth,height=5.5cm]{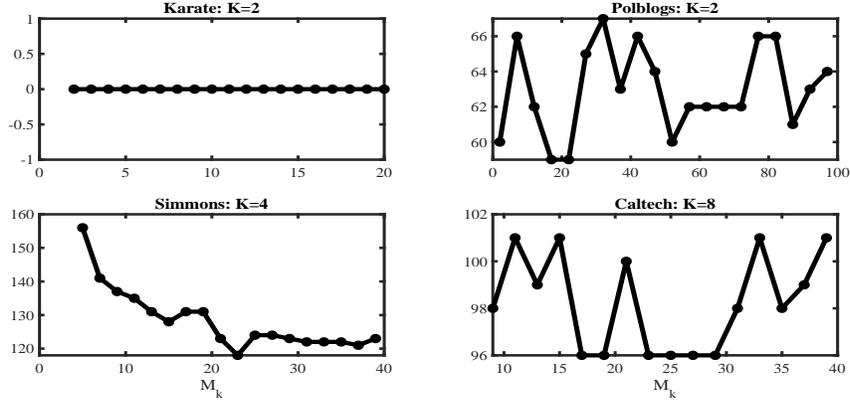}
	\caption{$\mathrm{PCC}^{*}$ is insensitive to $M_{k}$. For Karate club, $M_{k}\in \{2,3,4,5\ldots, 20\}$; for Political blogs, $M_{k}\in \{2,7,12,17,\ldots, 97\}$; for Simmons, $M_{k}\in\{5,7,9,11,\ldots, 39\}$; for Caltech, $M_{k}\in \{9,11,13,15,\ldots,39\}$. x-axis: $M_{k}$, y-axis: number errors.}\label{MkPCC}
\end{figure}

Figure \ref{MkPCC} records the relationship between $M_{k}$ and the number of errors for the four real-world datasets for $\mathrm{PCC}^{*}$. From Figure \ref{MkPCC} we can find that $\mathrm{PCC}^{*}$ is insensitive to $M_{k}$. Especially for Simmons and Caltech when $M_{k}\geq K+1$, the number of errors for both two datasets  decreases amazingly quickly. For Simmons, the number of errors of $\mathrm{PCC}^{*}$ can be smaller than 120, which is much smaller than the number of errors 243 of PCC; for Caltech, the number of errors can be as small as 96 which is much smaller than the number of errors 166 of PCC. This property of $\mathrm{PCC}^{*}$ suggests that we can apply more than $K$ eigenvectors and eigenvalues in the PCC approach for community detection and such procedure can give us more accurate detection results. For space saving, we leave the theoretical study on both $\mathrm{PCC}^{*}$ and $\mathrm{NPCC}^{*}$ as future work. And we also argue that whether there exists an optimal $M^{*}_{k}$ for $\mathrm{PCC}^{*}$ (as well as $\mathrm{NPCC}^{*}$) such that $\mathrm{PCC}^{*}$ based on $M^{*}_{k}$ outperforms all other $\mathrm{PCC}^{*}$ based on $M_{l}$, and we leave the study of this question as our future work.

\begin{figure}[H] \includegraphics[width=\textwidth,height=5.5cm]{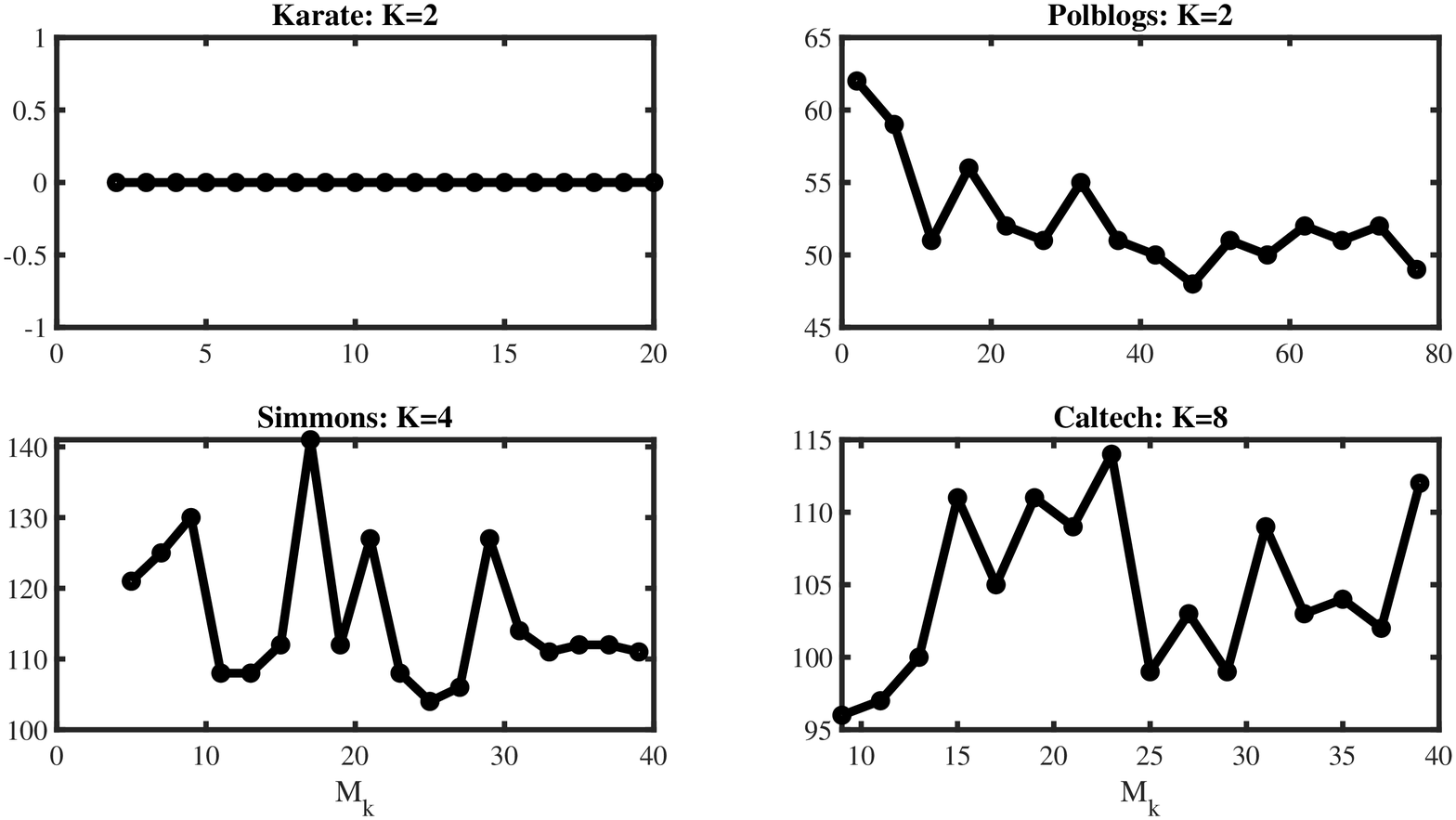}
	\caption{$\mathrm{NPCC}^{*}$ is insensitive to $M_{k}$. For Karate club, $M_{k}\in \{2,3,4,5\ldots, 20\}$; for Political blogs, $M_{k}\in \{2,7,12,17,\ldots, 77\}$; for Simmons, $M_{k}\in\{5,7,9,11,\ldots, 39\}$; for Caltech, $M_{k}\in \{9,11,13,15,\ldots,39\}$. x-axis: $M_{k}$, y-axis: number errors.}\label{MkNPCC}
\end{figure}

Figure \ref{MkNPCC} records the relationship between $M_{k}$  and the number of errors for the four datasets for $\mathrm{NPCC}^{*}$. Similarly, we can make a similar conclusion as $\mathrm{PCC}^{*}$.
 From Figure \ref{MkNPCC} we can find that $\mathrm{NPCC}^{*}$ is insensitive to $M_{k}$. For Karate club, the number error is always zero; for Political blog, the number error of $\mathrm{NPCC}^{*}$ reaches the minimum number error 48 when setting $M_{k}$ as 47;  for Simmons, the number error of $\mathrm{NPCC}^{*}$ reaches the minimum number error 104 (which is much smaller than the number error 225 of NPCC) when setting $M_{k}$ as 25; for Caltech, the number error of $\mathrm{NPCC}^{*}$ reaches the minimum number error 96 (which is much smaller than the number error 155 of NPCC) when setting $M_{k}$ as 5. This property of $\mathrm{NPCC}^{*}$ suggests us that we can apply more than $K$ eigenvectors and eigenvalues in NPCC approach for community detection and such procedure has no big influence on the community detection results. Meanwhile, by checking Figure \ref{MkNPCC}, we'd note that taking more than $K$ eigenvectors for clustering of NPCC when dealing with Simmons and Caltech can always have much smaller number errors compared with number errors of NPCC.

\begin{figure}[H] \includegraphics[width=\textwidth,height=5.5cm]{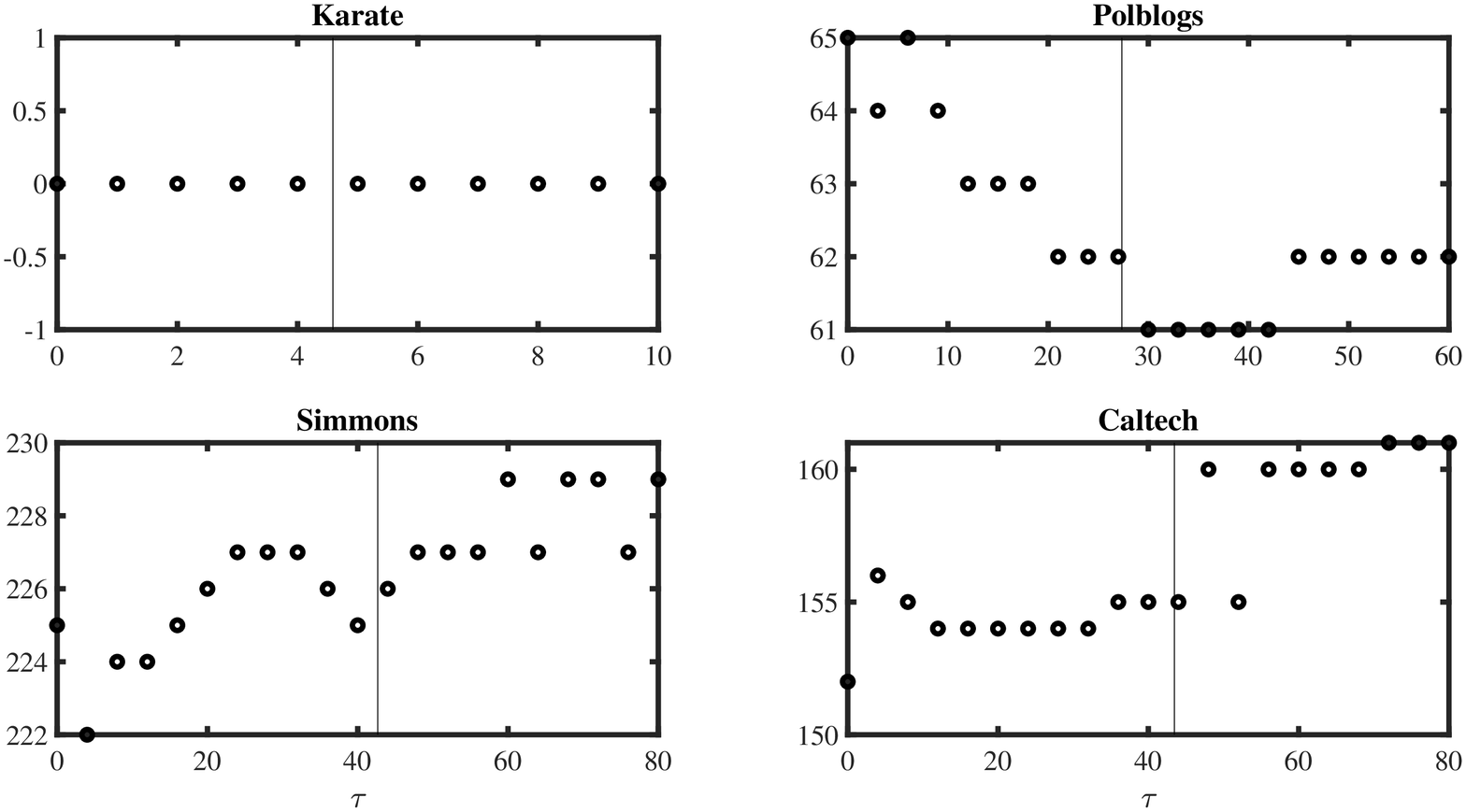}
	\caption{NPCC is insensitive to $\tau$. For Karate club, $\tau\in \{0,1,2,\ldots, 10\}$; for Political blogs, $\tau\in \{0,3,6,\ldots, 60\}$; for Simmons, $\tau\in\{0,4,8,\ldots, 80\}$; for Caltech, $\tau\in \{0,4,8,\ldots, 80\}$. x-axis: $\tau$, y-axis: number errors, vertical line: the default $\tau$ for each dataset.}\label{tauNPCC}
\end{figure}

 What's more, as shown by Figure \ref{tauNPCC} (which is obtained by setting $\tau$ ranging \footnote{Note that the default $\tau$ for Karate is 4.5882, for Polblogs is 27.3552; for Simmons is 42.6684, and for Caltech is 43.4644.} from certain intervals for a certain real dataset for NPCC method and then plot the respective number error), we can find that our NPCC approach inherits the nice property of RSC such that NPCC is insensitive to $\tau$ as long as $\tau > 0$. For example: for Karate network, when $\tau$ ranges in $\{0, 1, 2, \ldots, 10\}$, the number errors for NPCC on Karate are always 0; for Polblogs, when $\tau$ ranges in $\{0, 3, 6, \ldots, 60\}$, number errors for NPCC on Polblogs ranges around 62. Similar phenomenons occur for Simmons and Caltech, which support the conclusion \footnote{Note that since there is no full theoretical analysis that gives the theoretical bound of Hamming error rate for NPCC due to the fact that weyl's inequality fails to function on asymmetric matrices, we can not find the theoretical connections between $\tau$ and the theoretical bound of Hamming error rate for NPCC. This is the reason that we show that NPCC is insensitive to the choice of $\tau$ by numerical studies in this part. And we leave the theoretical study on the relationship between $\tau$ and the bound of Hamming error rate for NPCC for our future work.} that NPCC is insensitive to $\tau$ as long as $\tau > 0$.
\section{Discussion}\label{sec6}
In this paper, we design two new spectral methods PCC and  NPCC for community detection under DCSBM. NPCC is designed based on the PCC and the normalized Laplacian matrix $N^{L_{\tau}}$. Theoretical analysis shows that PCC holds consistent community detection under mild conditions, and NPCC returns perfect clustering under the ideal case. PCC is simple to analyse and  there is no tuning, threshold or regularizer parameters to be estimated. Numerical results show that PCC enjoys comparable performances with SCORE and OCCAM. Meanwhile, both synthetic and empirical results demonstrate that NPCC enjoys satisfactory performances compared with the state-of-the-art spectral clustering methods. What's more, since NPCC is designed based on PCC and RSC algorithms, numerical results show that NPCC provides a significant improvement compared with PCC and RSC. For the computing time when dealing with large networks, NPCC compares favorably with PCC, SCORE, OCCAM and RSC, and our NPCC is much faster than CMM since CMM is a modularity maximization method while NPCC is a spectral clustering method for community detection.

When dealing with weak signal networks, by considering one more eigenvectors for clustering, we have two refinements algorithms called PCC+ and NPCC+ designed based on PCC and NPCC, respectively. When dealing with two weak signal networks Simmons and Caltech, PCC+ provides significant improvement compared with PCC, and NPCC+ enjoys best performances with error rates 121/1137, and 96/590, respectively.

Meanwhile, NPCC owns two nice properties such that NPCC is insensitive to the number of eigenvectors for clustering, and NPCC is insensitive to the regularizer $\tau$ as long as $\tau\geq 0$. Since the theoretical guarantee about these two properties of NPCC is unknown at present, we leave it as future work. Due to the fact that the Weyl's inequality only works for symmetric matrices while $N^{L_{\tau}}$ and $N^{\mathscr{L}_{\tau}}$ are two asymmetric matrices, Weyl's inequality can not be applied to $N^{L_{\tau}}$ and $N^{\mathscr{L}_{\tau}}$ to find the bound of the difference of their eigenvalues. Therefore we can not finish the full theoretical analysis for NPCC at present and we leave it as future work.

We'd emphasize  that  our NPCC is totally designed based on an asymmetric matrix for community detection and it enjoys satisfactory numerical performances. Meanwhile, the work of PCC and NPCC can be extended to various problems, such as mixed membership community detection problem, weighted and directed network community detection problem, the problem of estimating the number of clusters $K$ based on the asymmetric matrix $N^{L_{\tau}}$ provided in this paper. We'll study these interesting problems in our future work.

\section*{Acknowledgements}

The authors would like to thank Assistant Professor Xin Tong, Thomson and Assistant Professor Wanjie Wang at National University of Singapore for many helpful discussions.

\bibliographystyle{Chicago}
\bibliography{refnpcc}

\begin{thebibliography}{29}
\expandafter\ifx\csname natexlab\endcsname\relax\def\natexlab#1{#1}\fi
\providecommand{\url}[1]{\texttt{#1}}
\providecommand{\href}[2]{#2}
\providecommand{\path}[1]{#1}
\providecommand{\DOIprefix}{doi:}
\providecommand{\ArXivprefix}{arXiv:}
\providecommand{\URLprefix}{URL: }
\providecommand{\Pubmedprefix}{pmid:}
\providecommand{\doi}[1]{\href{http://dx.doi.org/#1}{\path{#1}}}
\providecommand{\Pubmed}[1]{\href{pmid:#1}{\path{#1}}}
\providecommand{\bibinfo}[2]{#2}
\ifx\xfnm\relax \def\xfnm[#1]{\unskip,\space#1}\fi
\bibitem[{Adamic and Glance(2005)}]{Polblogs1}
\bibinfo{author}{Adamic, L.A.}, \bibinfo{author}{Glance, N.},
  \bibinfo{year}{2005}.
\newblock \bibinfo{title}{The political blogosphere and the 2004 us election:
  divided they blog} , \bibinfo{pages}{36--43}.
\bibitem[{{Amini} et~al.(2013){Amini}, {Chen}, {Bickel} and
  {Levina}}]{amini2013pseudo}
\bibinfo{author}{{Amini}, A.A.}, \bibinfo{author}{{Chen}, A.},
  \bibinfo{author}{{Bickel}, P.J.}, \bibinfo{author}{{Levina}, E.},
  \bibinfo{year}{2013}.
\newblock \bibinfo{title}{{Pseudo-likelihood methods for community detection in
  large sparse networks}}.
\newblock \bibinfo{journal}{Annals of Statistics} \bibinfo{volume}{41},
  \bibinfo{pages}{2097--2122}.
\bibitem[{Bickel and Chen(2009)}]{PJBAC}
\bibinfo{author}{Bickel, P.J.}, \bibinfo{author}{Chen, A.},
  \bibinfo{year}{2009}.
\newblock \bibinfo{title}{A nonparametric view of network models and
  newman--girvan and other modularities}.
\newblock \bibinfo{journal}{Proceedings of the National Academy of Sciences}
  \bibinfo{volume}{106}, \bibinfo{pages}{21068--21073}.
\bibitem[{Chaudhuri et~al.(2012)Chaudhuri, Chung and Tsiatas}]{RSCtau}
\bibinfo{author}{Chaudhuri, K.}, \bibinfo{author}{Chung, F.},
  \bibinfo{author}{Tsiatas, A.}, \bibinfo{year}{2012}.
\newblock \bibinfo{title}{Spectral clustering of graphs with general degrees in
  the extended planted partition model}.
\newblock \bibinfo{journal}{Journal of Machine Learning Research} ,
  \bibinfo{pages}{1--23}.
\bibitem[{{Chen} et~al.(2018){Chen}, {Li} and {Xu}}]{CMM}
\bibinfo{author}{{Chen}, Y.}, \bibinfo{author}{{Li}, X.},
  \bibinfo{author}{{Xu}, J.}, \bibinfo{year}{2018}.
\newblock \bibinfo{title}{Convexified modularity maximization for
  degree-corrected stochastic block models}.
\newblock \bibinfo{journal}{Annals of Statistics} \bibinfo{volume}{46},
  \bibinfo{pages}{1573--1602}.
\bibitem[{{Daudin} et~al.(2008){Daudin}, {Picard} and {Robin}}]{daudin2008a}
\bibinfo{author}{{Daudin}, J.J.}, \bibinfo{author}{{Picard}, F.},
  \bibinfo{author}{{Robin}, S.}, \bibinfo{year}{2008}.
\newblock \bibinfo{title}{A mixture model for random graphs}.
\newblock \bibinfo{journal}{Statistics and Computing} \bibinfo{volume}{18},
  \bibinfo{pages}{173--183}.
\bibitem[{Fortunato and Hric(2016)}]{fortunato2016community}
\bibinfo{author}{Fortunato, S.}, \bibinfo{author}{Hric, D.},
  \bibinfo{year}{2016}.
\newblock \bibinfo{title}{Community detection in networks: A user guide}.
\newblock \bibinfo{journal}{Physics Reports} \bibinfo{volume}{659},
  \bibinfo{pages}{1--44}.
\bibitem[{{Holland} et~al.(1983){Holland}, {Laskey} and {Leinhardt}}]{SBM}
\bibinfo{author}{{Holland}, P.W.}, \bibinfo{author}{{Laskey}, K.B.},
  \bibinfo{author}{{Leinhardt}, S.}, \bibinfo{year}{1983}.
\newblock \bibinfo{title}{Stochastic blockmodels: First steps}.
\newblock \bibinfo{journal}{Social Networks} \bibinfo{volume}{5},
  \bibinfo{pages}{109--137}.
\bibitem[{{Jin}(2015)}]{SCORE}
\bibinfo{author}{{Jin}, J.}, \bibinfo{year}{2015}.
\newblock \bibinfo{title}{{Fast community detection by SCORE}}.
\newblock \bibinfo{journal}{Annals of Statistics} \bibinfo{volume}{43},
  \bibinfo{pages}{57--89}.
\bibitem[{{Jin} et~al.(2018){Jin}, {Ke} and {Luo}}]{SCORE+}
\bibinfo{author}{{Jin}, J.}, \bibinfo{author}{{Ke}, Z.T.},
  \bibinfo{author}{{Luo}, S.}, \bibinfo{year}{2018}.
\newblock \bibinfo{title}{{SCORE+ for network community detection}}.
\newblock \bibinfo{journal}{arXiv preprint arXiv:1811.05927} .
\bibitem[{{Karrer} and {Newman}(2011)}]{DCSBM}
\bibinfo{author}{{Karrer}, B.}, \bibinfo{author}{{Newman}, M.E.J.},
  \bibinfo{year}{2011}.
\newblock \bibinfo{title}{Stochastic blockmodels and community structure in
  networks}.
\newblock \bibinfo{journal}{Physical Review E} \bibinfo{volume}{83},
  \bibinfo{pages}{16107}.
\bibitem[{Liu et~al.(2018)Liu, Choi, Xie and Roeder}]{liu2018global}
\bibinfo{author}{Liu, F.}, \bibinfo{author}{Choi, D.S.}, \bibinfo{author}{Xie,
  L.}, \bibinfo{author}{Roeder, K.}, \bibinfo{year}{2018}.
\newblock \bibinfo{title}{Global spectral clustering in dynamic networks}.
\newblock \bibinfo{journal}{Proceedings of the National Academy of Sciences of
  the United States of America} \bibinfo{volume}{115},
  \bibinfo{pages}{927--932}.
\bibitem[{{Luxburg}(2007)}]{Tutorial}
\bibinfo{author}{{Luxburg}, U.}, \bibinfo{year}{2007}.
\newblock \bibinfo{title}{A tutorial on spectral clustering}.
\newblock \bibinfo{journal}{Statistics and Computing} \bibinfo{volume}{17},
  \bibinfo{pages}{395--416}.
\bibitem[{Newman(2006)}]{MN2006}
\bibinfo{author}{Newman, M.E.}, \bibinfo{year}{2006}.
\newblock \bibinfo{title}{Modularity and community structure in networks}.
\newblock \bibinfo{journal}{Proceedings of the national academy of sciences}
  \bibinfo{volume}{103}, \bibinfo{pages}{8577--8582}.
\bibitem[{Newman(2004)}]{newman2004detecting}
\bibinfo{author}{Newman, M.E.J.}, \bibinfo{year}{2004}.
\newblock \bibinfo{title}{Detecting community structure in networks}.
\newblock \bibinfo{journal}{European Physical Journal B} \bibinfo{volume}{38},
  \bibinfo{pages}{321--330}.
\bibitem[{Ng et~al.(2001)Ng, Jordan and Weiss}]{ng2001on}
\bibinfo{author}{Ng, A.Y.}, \bibinfo{author}{Jordan, M.I.},
  \bibinfo{author}{Weiss, Y.}, \bibinfo{year}{2001}.
\newblock \bibinfo{title}{On spectral clustering: Analysis and an algorithm}.
\newblock \bibinfo{journal}{Neural Information Processing Systems} ,
  \bibinfo{pages}{849--856}.
\bibitem[{Nguyen et~al.(2014)Nguyen, Dinh, Shen and Thai}]{nguyen2014dynamic}
\bibinfo{author}{Nguyen, N.P.}, \bibinfo{author}{Dinh, T.N.},
  \bibinfo{author}{Shen, Y.}, \bibinfo{author}{Thai, M.T.},
  \bibinfo{year}{2014}.
\newblock \bibinfo{title}{Dynamic social community detection and its
  applications.}
\newblock \bibinfo{journal}{PLOS ONE} \bibinfo{volume}{9}.
\bibitem[{{Nowicki} and {Snijders}(2001)}]{nowicki2001estimation}
\bibinfo{author}{{Nowicki}, K.}, \bibinfo{author}{{Snijders}, T.A.B.},
  \bibinfo{year}{2001}.
\newblock \bibinfo{title}{Estimation and prediction for stochastic
  blockstructures}.
\newblock \bibinfo{journal}{Journal of the American Statistical Association}
  \bibinfo{volume}{96}, \bibinfo{pages}{1077--1087}.
\bibitem[{{Qin} and {Rohe}(2013)}]{RSC}
\bibinfo{author}{{Qin}, T.}, \bibinfo{author}{{Rohe}, K.},
  \bibinfo{year}{2013}.
\newblock \bibinfo{title}{{Regularized spectral clustering under the
  degree-corrected stochastic blockmodel}}, in: \bibinfo{booktitle}{Advances in
  Neural Information Processing Systems 26}, pp. \bibinfo{pages}{3120--3128}.
\bibitem[{{Rohe} et~al.(2011){Rohe}, {Chatterjee} and {Yu}}]{rohe2011spectral}
\bibinfo{author}{{Rohe}, K.}, \bibinfo{author}{{Chatterjee}, S.},
  \bibinfo{author}{{Yu}, B.}, \bibinfo{year}{2011}.
\newblock \bibinfo{title}{Spectral clustering and the high-dimensional
  stochastic blockmodel}.
\newblock \bibinfo{journal}{Annals of Statistics} \bibinfo{volume}{39},
  \bibinfo{pages}{1878--1915}.
\bibitem[{Rohe et~al.(2011)Rohe, Chatterjee and Yu}]{nPCA}
\bibinfo{author}{Rohe, K.}, \bibinfo{author}{Chatterjee, S.},
  \bibinfo{author}{Yu, B.}, \bibinfo{year}{2011}.
\newblock \bibinfo{title}{Spectral clustering and the high-dimensional
  stochastic blockmodel}.
\newblock \bibinfo{journal}{The Annals of Statistics} \bibinfo{volume}{39},
  \bibinfo{pages}{1878--1915}.
\bibitem[{{Snijders} and {Nowicki}(1997)}]{snijders1997estimation}
\bibinfo{author}{{Snijders}, T.A.B.}, \bibinfo{author}{{Nowicki}, K.},
  \bibinfo{year}{1997}.
\newblock \bibinfo{title}{{Estimation and prediction for stochastic blockmodels
  for graphs with latent block structur}}.
\newblock \bibinfo{journal}{Journal of Classification} \bibinfo{volume}{14},
  \bibinfo{pages}{75--100}.
\bibitem[{{Sylvester}(1852)}]{sylvester1852xix}
\bibinfo{author}{{Sylvester}, J.}, \bibinfo{year}{1852}.
\newblock \bibinfo{title}{{XIX. A demonstration of the theorem that every
  homogeneous quadratic polynomial is reducible by real orthogonal
  substitutions to the form of a sum of positive and negative squares}}.
\newblock \bibinfo{journal}{Philosophical Magazine Series 1}
  \bibinfo{volume}{4}, \bibinfo{pages}{138--142}.
\bibitem[{Tomasoni et~al.(2020)Tomasoni, Gómez, Crawford, Zhang, Choobdar,
  Marbach and Bergmann}]{tomasoni2020monet}
\bibinfo{author}{Tomasoni, M.}, \bibinfo{author}{Gómez, S.},
  \bibinfo{author}{Crawford, J.}, \bibinfo{author}{Zhang, W.},
  \bibinfo{author}{Choobdar, S.}, \bibinfo{author}{Marbach, D.},
  \bibinfo{author}{Bergmann, S.}, \bibinfo{year}{2020}.
\newblock \bibinfo{title}{Monet: a toolbox integrating top-performing methods
  for network modularization}.
\newblock \bibinfo{journal}{Bioinformatics} \bibinfo{volume}{36},
  \bibinfo{pages}{3920--3921}.
\bibitem[{Traud et~al.(2011)Traud, Kelsic, Mucha and
  Porter}]{traud2011comparing}
\bibinfo{author}{Traud, A.L.}, \bibinfo{author}{Kelsic, E.D.},
  \bibinfo{author}{Mucha, P.J.}, \bibinfo{author}{Porter, M.A.},
  \bibinfo{year}{2011}.
\newblock \bibinfo{title}{Comparing community structure to characteristics in
  online collegiate social networks}.
\newblock \bibinfo{journal}{Siam Review} \bibinfo{volume}{53},
  \bibinfo{pages}{526--543}.
\bibitem[{Traud et~al.(2012)Traud, Mucha and Porter}]{traud2012social}
\bibinfo{author}{Traud, A.L.}, \bibinfo{author}{Mucha, P.J.},
  \bibinfo{author}{Porter, M.A.}, \bibinfo{year}{2012}.
\newblock \bibinfo{title}{Social structure of facebook networks}.
\newblock \bibinfo{journal}{Physica A-statistical Mechanics and Its
  Applications} \bibinfo{volume}{391}, \bibinfo{pages}{4165--4180}.
\bibitem[{Weyl(1912)}]{Weyl}
\bibinfo{author}{Weyl, H.}, \bibinfo{year}{1912}.
\newblock \bibinfo{title}{Das asymptotische verteilungsgesetz der eigenwerte
  linearer partieller differentialgleichungen (mit einer anwendung auf die
  theorie der hohlraumstrahlung)}.
\newblock \bibinfo{journal}{Mathematische Annalen} \bibinfo{volume}{71},
  \bibinfo{pages}{441--479}.
\bibitem[{Zachary(1977)}]{karate}
\bibinfo{author}{Zachary, W.W.}, \bibinfo{year}{1977}.
\newblock \bibinfo{title}{An information flow model for conflict and fission in
  small groups}.
\newblock \bibinfo{journal}{Journal of anthropological research}
  \bibinfo{volume}{33}, \bibinfo{pages}{452--473}.
\bibitem[{{Zhang} et~al.(2020){Zhang}, {Levina} and {Zhu}}]{OCCAM}
\bibinfo{author}{{Zhang}, Y.}, \bibinfo{author}{{Levina}, E.},
  \bibinfo{author}{{Zhu}, J.}, \bibinfo{year}{2020}.
\newblock \bibinfo{title}{{Detecting overlapping communities in networks using
  spectral methods}}.
\newblock \bibinfo{journal}{SIAM Journal on Mathematics of Data Science}
  \bibinfo{volume}{2}, \bibinfo{pages}{265--283}.

\end{thebibliography}

\end{document}